\documentclass[letterpaper]{article} 
\usepackage{aaai2027}  
\usepackage[hyphens]{url}  
\usepackage{graphicx} 
\usepackage{natbib}  
\usepackage{caption} 
\usepackage{algorithm}
\usepackage{amsmath}
\usepackage{booktabs}
\usepackage{tabularx}
\usepackage{array}
\usepackage[most]{tcolorbox}
\usepackage{algpseudocode}
\usepackage[table]{xcolor}
\definecolor{blockhdr}{RGB}{230,238,246}
\usepackage{newfloat}
\usepackage{listings}
\usepackage{xspace}
\usepackage[most]{tcolorbox}

\newif\ifshowcomments

\showcommentsfalse    

\newcommand{\yaoqi}[1]{%
  \ifshowcomments
    \textcolor{blue}{(#1 --yaoqi)}\xspace
  \fi
}
\newcommand{\beidi}[1]{%
  \ifshowcomments
    \textcolor{red}{(#1 --beidi)}\xspace
  \fi
}

\usepackage{xspace}
\newcommand{\MESA}{\texttt{MESA}\xspace}
\newtcolorbox{axisbox}[1]{
colback=gray!4, colframe=gray!55, boxrule=0.6pt, arc=2pt,
left=6pt, right=6pt, top=4pt, bottom=4pt,
breakable, enhanced,
fonttitle=\bfseries\small, coltitle=black,
title={#1}
}

\DeclareCaptionStyle{ruled}{labelfont=normalfont,labelsep=colon,strut=off} 
\floatstyle{ruled}
\newfloat{listing}{tb}{lst}{}
\floatname{listing}{Listing}

\usepackage{booktabs}

\usepackage{booktabs} 
\usepackage{multirow} 
\usepackage{amssymb}
\usepackage{threeparttable} 
\usepackage[table]{xcolor}
\title{\MESA: Task-Adaptive Multi-Structure Evidence \\Selection for Long-Horizon Agent Memory}
\author{ 
    Beidi Zhao\textsuperscript{\rm 1}\footnote{The work is completed during internship at Microsoft Research Asia.}, 
    Yaoqi Chen\textsuperscript{\rm 2,3},
    Yuru Feng\textsuperscript{\rm 2,4},
    Menghao Li\textsuperscript{\rm 2}, 
    Qianxi Zhang\textsuperscript{\rm 2}, 
    Baotong Lu\textsuperscript{\rm 2}, 
    Jianan Lu\textsuperscript{\rm 2}, 
    Zhirui Wang\textsuperscript{\rm 2}, 
    Xinjiang Wang\textsuperscript{\rm 2}, 
    Shusen Xu\textsuperscript{\rm 2}, 
    Zengzhong Li\textsuperscript{\rm 2}, 
    Xiaoxiao Li\textsuperscript{\rm 1}\corresponding, 
    Qi Chen\textsuperscript{\rm 2}\corresponding
}
\affiliations{
    \textsuperscript{\rm 1}University of British Columbia, \textsuperscript{\rm 2}Microsoft, \\\textsuperscript{\rm 3}University of Science and Technology of China, 
    \textsuperscript{\rm 4}University of California, San Diego\\

}

\begin{document}

\maketitle

\begin{abstract}
Long-horizon agents accumulate trajectories spanning hundreds of interleaved reasoning, action, and observation steps, where answering a query may depend on evidence buried far back in the history. External memory stores such trajectories as structured representations, yet each structure provides a distinct and incomplete view. Existing multi-memory systems either read a fixed set of structures for every query, inflating context and introducing noise, or route each query to a single structure, preventing the composition of complementary evidence. 
A controlled analysis on AMA-Bench shows that the optimal memory configuration is typically neither a single structure nor the full union, but a tailored composition of multiple structural memories that varies with query and task demands. Motivated by these findings, we formulate \emph{structure-level dynamic selection}: selecting and fusing a query-adaptive subset from a library of specialized memory structures. 
We propose \MESA (a \underline{\texttt{M}}ulti-structure \underline{\texttt{E}}vidence \underline{\texttt{S}}election framework for long-horizon \underline{\texttt{A}}gent), which builds five complementary structure views of each trajectory and learns from end-to-end answer-level feedback to select and fuse a query-specific subset for a frozen answer model. To learn under this weak supervision, \MESA employs harness optimization with prior-guided search and UCB-guided scheduling to balance exploration and exploitation. On AMA-Bench, \MESA outperforms the strongest baseline by 8.5\% while using 41\% fewer evidence tokens than the all-structure alternative.

\end{abstract}


\section{Introduction}
Long-horizon agents accumulate trajectories that span hundreds of steps, where a later decision can depend on evidence
produced much earlier \cite{yao2023react,wang2023voyager,zhao2026ama}. Unlike documents or dialogues, these
trajectories interleave reasoning, actions, observations, and tool outputs, and they carry temporal and causal
dependencies \cite{zhao2026ama}. Feeding the full trajectory to a long-context LLM is expensive and still does not
reliably surface buried evidence, even with million-token windows \cite{team2024gemini,liu2024lost}. Truncation and
compression cut costs but can drop details that matter later \cite{jiang2023llmlingua}. External memory addresses this problem by transforming the history into persistent representations and retrieving a compact body of evidence for each query \cite{packer2023memgpt,park2023generative,zhong2024memorybank}. Accordingly, the central question is how execution histories should be organized and dynamically consulted.

\begin{figure}[!t]
    \centering
    \includegraphics[width=\linewidth]{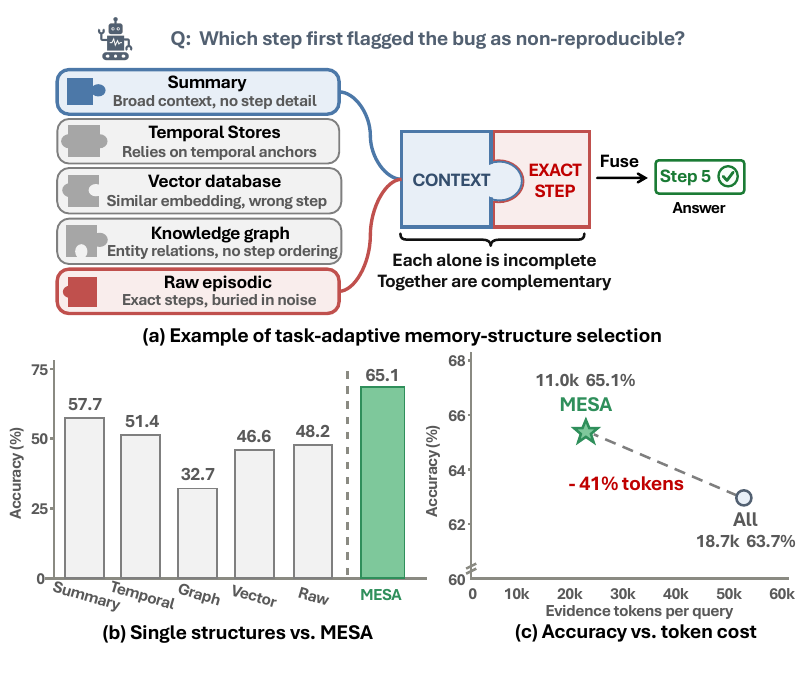}
    \caption{
    Overview and motivation of \MESA.
    (a) An SWE long-horizon agent execution example. Each single memory structure provides a distinct view of evidence. \MESA dynamically selects and fuses a subset of complementary structural memories to arrive at the correct answer.
    (b) Driven by adaptive memory-structure selection, \MESA outperforms all single-structure baselines.
    (c) \MESA achieves higher accuracy while consuming 41\% fewer tokens compared to the all-structure alternative.
    }
    \label{fig:teaser}
\end{figure}

Existing agent-memory systems rely on distinct structural abstractions to manage execution histories, yet each abstraction exhibits inherent limitations when handling the heterogeneity of long-horizon agent applications \cite{zhao2026ama}. As illustrated in Fig.~\ref{fig:teaser}(a), text summaries \cite{zhong2024memorybank} capture high-level context but lack deterministic execution details; temporal stores \cite{sen2026chronos} preserve chronological order but may struggle when queries lack explicit temporal anchors in the query; vector databases \cite{lewis2020retrieval,karpukhin2020dense} surface semantically near fragments that can be task-irrelevant while discarding temporal-causal order; knowledge graphs \cite{yang2603plugmem,gutierrez2024hipporag} facilitate entity traversal but lack step-level indexing; and raw episodic traces \cite{zheng2024synapse} retain exact action steps but bury critical clues under heavy noise.

To bridge individual limitations, recent hybrid memory systems combine these structures, but their access strategies remain mismatched with
query-specific demands. At one extreme, \textbf{eager all-in fusion} methods uniformly query a static, exhaustive set of memory views for every query \cite{latimer2025hindsight,su2026s3mem}. While maximizing recall, this indiscriminately bloats context windows and introduces distracting noise from unneeded structural memories \cite{ha2026memguard,narayana2026diagnosing}. At the other extreme, \textbf{single-structure routing} methods select a single best-suited structure per query \cite{lu2026fluxmem,bai2026learning}. However, this strictly limits expressiveness when a query demands composing complementary evidence across representations (e.g., grounding a high-level summary within exact episodic steps, as in Fig.~\ref{fig:teaser}(a)). Consequently, no existing approach dynamically selects and composes a query-adaptive subset from a library of specialized memory structures.

To formalize this problem, we define a \emph{memory structure} as a unified pair comprising a structural representation and its dedicated access interface. Guided by agent memory taxonomies \cite{wu2025human,luo2026storage}, we instantiate five representative structures: text summaries, temporal stores, knowledge graphs, vector databases, and raw episodic traces. To quantify the intrinsic value of structure composition, we conduct a controlled study sweeping (details in Sec.~\ref{sec:analysis}) over all non-empty structure subsets across diverse long-horizon tasks. 
This controlled analysis reveals two key findings:
(1) \emph{Intermediate subsets are usually preferable:}
across most evaluated categories, the best-performing
composition is neither a single structure nor the full union,
but a subset of complementary structures; and
(2) \emph{No fixed composition is universally optimal over tasks:}
the winning subset varies across domains and memory
capabilities, rather than following a single global preference.
These findings motivate a new problem formulation: \emph{structure-level dynamic selection}, learning an executable policy to dynamically select and fuse query-dependent structural memories.

However, realizing an effective structure-level selection policy entails two fundamental challenges. First, selecting dynamic memory subsets is inherently a \textbf{combinatorial utility and redundancy dilemma}. Complementary views can improve evidence coverage, whereas redundant or mismatched views introduce distracting context and additional cost. The selector must therefore balance evidence recall, utility and context budget.
Second, policy learning faces \textbf{credit assignment under weak supervision}. Ground-truth subsets and per-structure utility labels are unavailable. The policy must instead be learned from sparse, end-to-end answer feedback, making individual selection decisions difficult to evaluate.

To tackle these challenges, we propose \MESA (\underline{\texttt{M}}ulti-structure \underline{\texttt{E}}vidence \underline{\texttt{S}}election for long-horizon \underline{\texttt{A}}gent). \MESA maintains the five independent memory structures: text summary, temporal store, knowledge graph, vector database, and raw episodic trace. Each with its own builder and retrieval interface. For each query, its selector dynamically choose query-dependent memory structures. Evidence from the selected structures are composed for a frozen answer-generating LLM. To optimize the selection policy under the weak supervision, \MESA restricts candidate selection to prior directions and balances the exploration-exploitation. 

We summarize our contributions: (1) \textbf{Problem formulation and empirical analysis.}
    We formulate structure-level dynamic selection for agent memory and show through an exhaustive subset sweep that intermediate subsets typically outperform the single-structure and all-structure alternatives, while the best composition varies across task categories. 
    (2) \textbf{The \MESA framework.}
    We introduce a multi-structure memory framework that learns query-adaptive structure selection policy from sparse answer-level feedback through prior-guided harness optimization.
    (3) \textbf{Empirical results.}
    \MESA outperforms the strongest baseline by 8.5\% on AMA-Bench. Results on LoCoMo extend the framework to conversational memory. Extensive ablations further validate the effectiveness of the learned task-adaptive selection.


\begin{figure}[!t]
    \centering
    \includegraphics[width=\columnwidth]
    {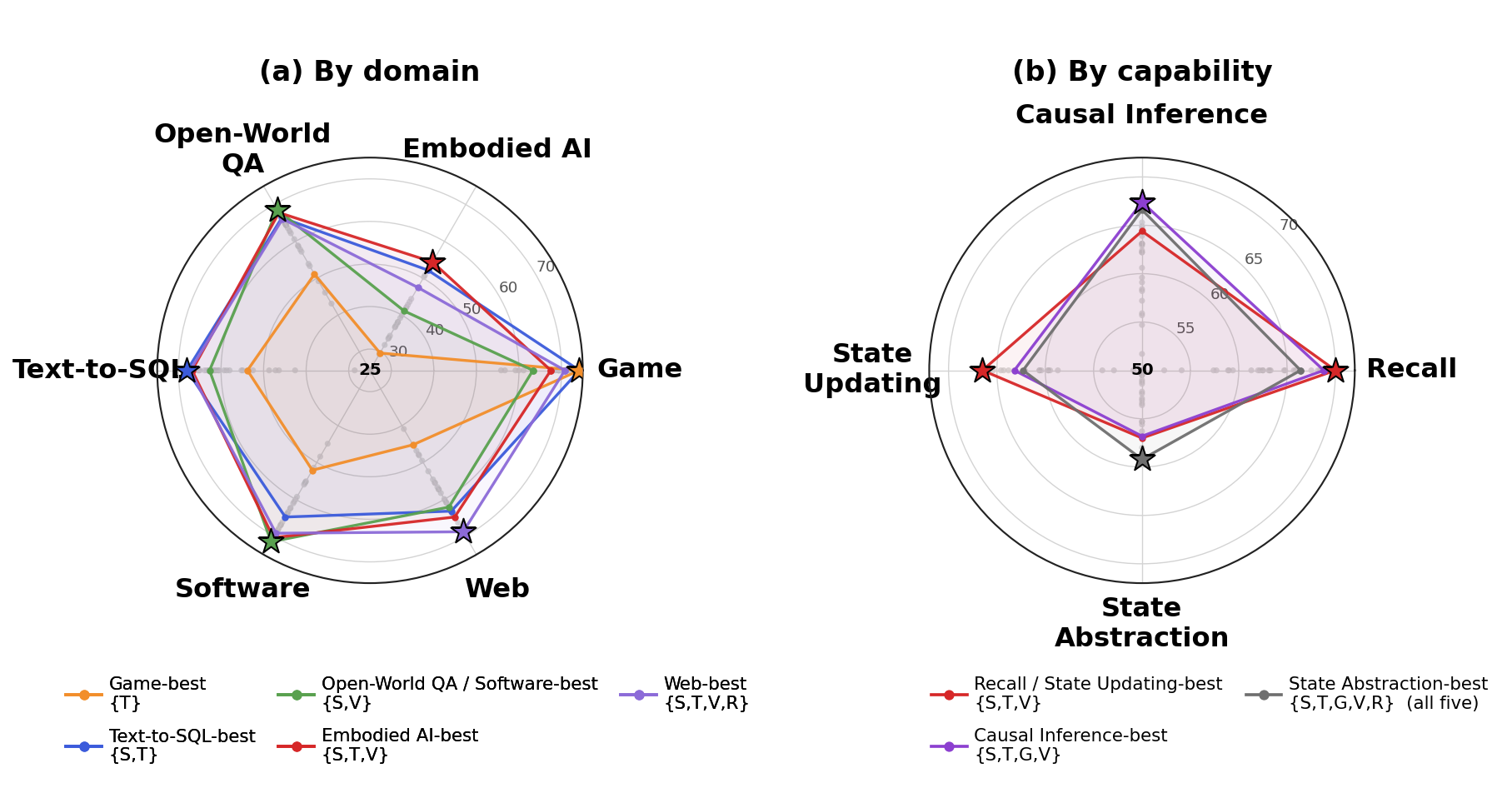}
    \caption{The best memory-structure combination is task-dependent. We sweep non-empty subsets of the five structures (Summary-S, Temporal-T, Graph-G, Vector-V, Raw-R) on AMA-Bench and report judge accuracy (\%), split (a) by domain and (b) by capability. Colored lines trace only the per-axis winners, each marked with a star. The winning subset changes on different tasks. Grey dots show all other combinations' result. 
    \beidi{Reviewer will automatically think of 'oracle', explain the gap and difficulty with case. We can explain in experiment or discussion}
    }
    \label{fig:motivating-analysis}
\end{figure}

\section{Related Work}
\subsection{Memory for LLM Agents}
Equipping LLM agents with memory beyond a fixed context window has been approached from several angles. MemGPT \cite{packer2023memgpt} manages the working context itself, paging information in and out like an operating system. 
A reflection-based line distills raw interaction into reusable knowledge: Generative Agents' retrieve–reflect memory \cite{park2023generative} and Reflexion's verbal
self-feedback across trials \cite{shinn2023reflexion}, while embodied agents such as Voyager accumulate a growing library of skills mined from long trajectories \cite{wang2023voyager},
underpinning agents that reuse distilled experience to act in multi-step environments \cite{yang2023auto}. A parallel line builds long-term memory for cross-session dialogue and multi-hop QA, and
several of these systems already maintain more than one index over the same history, pairing a graph store with dense or timestamped retrieval
\cite{zhong2024memorybank,chhikara2025mem0,gutierrez2024hipporag,maharana2024locomo,wu2024longmemeval}. Even where multiple indices coexist, however, each system reads them through a fixed
interface; how an agent should select among heterogeneous evidence forms per query, as long-horizon multi-task settings demand, is left unaddressed.

\subsection{Multi-Structure Memory for LLM Agents}
  A growing body of work organizes agent memory into explicit structures, but differs from ours along three axes: how memory is organized (by semantic role or a single unified schema versus by access
  pattern), how it is read (a fixed policy or a single structure versus a query-specific complementary subset), and what the read returns (broad or single-source evidence versus a fused subset of
  heterogeneous structures). One line reads its structures through a fixed policy: Hindsight \cite{latimer2025hindsight} partitions memory into four networks (world facts, experiences, opinions, and observations) but answers
  every query with the same four-way parallel retrieval fused by reciprocal rank fusion, and SEEM \cite{lu2026structured} combines a graph layer and an episodic layer through a deterministic
  pipeline; the read never adapts to what a query needs. A second line makes selection adaptive but returns a single winner: StructRAG \cite{li2025structrag} reconstructs documents into the single
  best-fit format per query, Learning-to-Route dispatches each query to one external source \cite{bai2026learning}, and FluxMem \cite{lu2026fluxmem} learns a context-aware selector that assigns
  each memory unit one structure among linear, graph, and hierarchical forms, supervised offline on conversational data. A third line, closest to ours, also selects compact
  evidence: S3Mem \cite{su2026s3mem} writes heterogeneous histories into a single unified scene–event schema and routes a small evidence pack to the reader, but its selection operates within one
  representation rather than across structures with distinct access patterns. \MESA differs on all three axes: it keeps heterogeneous structures distinguished, reads a query-specific complementary subset rather than one structure or the full union, and refines the selection policy through answer-level optimization.


\section{Analysis of Memory Structure Selection}
\label{sec:analysis}

\subsection{Memory Structure and Action Space}
A long interaction history admits several forms of storage and access. We use \emph{memory structure} to denote a memory representation together with its access mechanism, which jointly determine what information is retained and what evidence can be retrieved. Given a history $\tau_i$,
we construct $K$ memory structures:
\begin{equation}
\mathcal{M}_i
=
\big\{
M_i^{(k)}=B_k(\tau_i)
\big\}_{k=1}^{K},
\end{equation}
where $B_k$ is the builder for structure $k$, and $\Gamma_k$ is its fixed access mechanism. We instantiate $K=5$ representative structures widely used in agent-memory systems \cite{wu2025human,luo2026storage}: a compressed summary, a temporal key--value store, a relational graph, a dense vector index, and sparse retrieval over raw episodic traces. For a fair evaluation, we follow representative prior works~\cite{zhao2026ama,gutierrez2024hipporag} for the memory schema and construction of (see Appendix).

Reading memory requires choosing which structures to access. We
represent a selection as a non-empty binary vector $z\in\{0,1\}^{K}$, where $z^{(k)}=1$ indicates that structure $k$ is selected. The action space is
\begin{equation}
\mathcal{Z}
=
\left\{
z\in\{0,1\}^{K}
\mid
\lVert z\rVert_0\geq1
\right\},
\end{equation}
which contains $2^5-1=31$ non-empty compositions for $K=5$. We next evaluate these compositions to determine whether a single fixed subset is sufficient across tasks.

\subsection{Controlled Composition Analysis}
We evaluate all non-empty subsets of the five memory structures on AMA-Bench~\citep{zhao2026ama}. This controlled sweep reveals two observations.

\noindent\textbf{Intermediate subsets are usually preferable.}
Across the domains and QA capabilities in Fig.~\ref{fig:motivating-analysis}, most categories are best served by subsets containing multiple structures, indicating that they benefit from complementary evidence. However, adding every available structure is generally unnecessary and can introduce redundant or distracting context. The appropriate selection granularity therefore lies between route-to-one and read-all.

\noindent\textbf{No fixed composition is universally optimal over tasks:}
Fig.~\ref{fig:motivating-analysis} also shows that a composition
performing well for one domain or QA type may be suboptimal for another. No single fixed subset consistently provides the most useful evidence across tasks. The selected subset should therefore adapt to the current query rather than remain fixed.

Together, these findings motivate a selector $S_\rho(q_i,c_i)$ that predicts a non-empty subset $z_i\in\mathcal{Z}$ from the query and its observable context. The selector should exploit complementary structures when needed while avoiding the redundant evidence cost of the full composition. The next section formalizes this selector and describes how \MESA learns it from answer-level training feedback.

\begin{figure*}[!t]
    \centering
    \includegraphics[width=\textwidth]
    {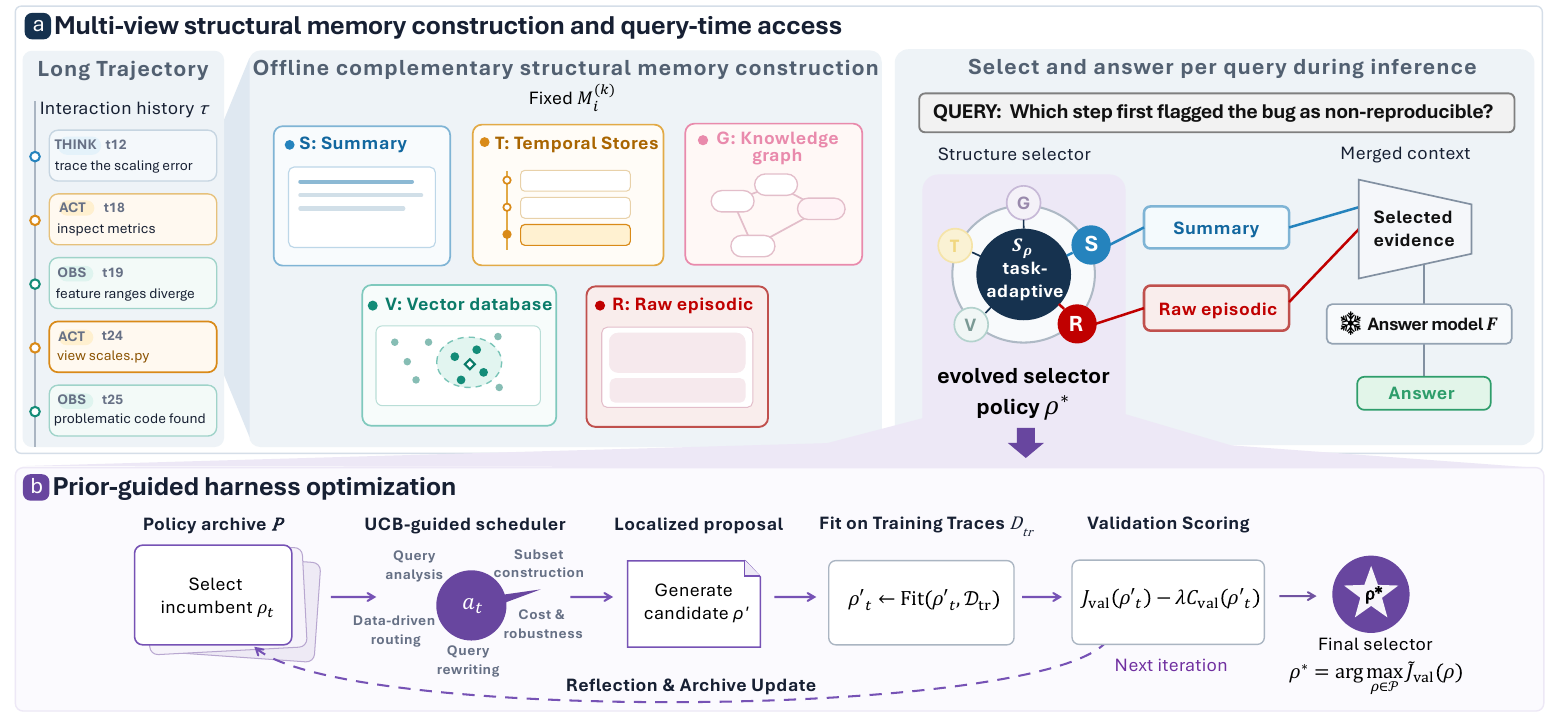}
    \caption{
Overview of \textsc{\MESA}.
(a) A long interaction trajectory is converted offline into five complementary memory structures. At inference, the learned selector $\rho^*$ chooses a query-dependent subset, retrieves and merges its evidence, and passes it to a
frozen answer model. (b) Prior-guided harness optimization evolves only the selector under prior directions and using UCB to balance exploration and exploitation. Localized proposals are evaluated using a performance--cost objective, and reflections update the policy archive. The final policy is selected on
validation data. 
}
    \label{fig:main}
\end{figure*}

\section{Task-Adaptive Memory Structure Selection}
\label{sec:method}

As illustrated in Fig.~\ref{fig:main}, \MESA learns an executable policy that selects a query-specific subset of five complementary memory structures, while keeping all other components fixed. We first formulate the selection problem and then describe the prior-guided policy optimization and evaluation protocol.


\subsection{Problem Formulation}

Given the memory structures
$\mathcal{M}_i=\{M_i^{(k)}\}_{k=1}^{K}$ constructed from history
$\tau_i$, the structure selector takes a query $q_i$ and a compact description $c_i$ of its interaction context and outputs a non-empty selection:
\begin{equation}
z_i=S_\rho(q_i,c_i),
\qquad
z_i\in\mathcal{Z},
\end{equation}
where $\rho$ is an executable selection policy and $\mathcal{Z}$ is the action space. The context $c_i$ may contain task, domain, and QA-type information, but excludes the gold
answer, evaluation score, and other outcome-revealing variables.

\MESA invokes the access mechanisms selected by $z_i$ and composes their retrieved evidence:
\begin{equation}
E_i(z_i)
=
\mathrm{Compose}
\left(
\left\{
\Gamma_k(q_i,M_i^{(k)})
\mid
z_i^{(k)}=1
\right\}
\right).
\end{equation}
A frozen answer model $F$ then performs inference over the selected evidence:
\begin{equation}
\hat{y}_i(z_i)
=
F\left(q_i,E_i(z_i)\right),
\end{equation}
and the resulting utility is
\begin{equation}
r_i(z_i)
=
\mathrm{Eval}\left(\hat{y}_i(z_i),y_i\right).
\end{equation}

We split the dataset into training, validation and test sets
$\mathcal{D}_{\mathrm{tr}}$,  $\mathcal{D}_{\mathrm{val}}$ and $\mathcal{D}_{\mathrm{test}}$. For a split $\mathcal{D}$, the evaluation score of policy $\rho$ is
\begin{equation}
J_{\mathcal{D}}(\rho)
=
\frac{1}{|\mathcal{D}|}
\sum_{i\in\mathcal{D}}
r_i\left(S_\rho(q_i,c_i)\right).
\end{equation}
Let $C_{\mathcal{D}}(\rho)$ denote the mean number of evidence tokens
retrieved from the selected structures. Candidate policies are compared
using the regularized validation objective
\begin{equation}
\widetilde{J}_{\mathrm{val}}(\rho)
=
J_{\mathrm{val}}(\rho)
-
\lambda C_{\mathrm{val}}(\rho),
\label{eq:selection-objective}
\end{equation}
where $\lambda$ controls the trade-off between answer accuracy and evidence cost.

\subsection{Prior-Guided Harness Optimization}
Feedback-driven optimizers can search over textual prompts
\cite{agrawal2025gepa} or complete executable harnesses
\cite{lee2026meta}, but such broad spaces are poorly directed
under the coarse, answer-level feedback available here
\cite{pan2026m,liu2026synthesizing,wu2026bayesian}. Since \MESA
keeps the builders, retrievers, composer, and answer model fixed, we
restrict optimization to a small set of \emph{prior directions}
$\mathcal{A} = \{a_1, \dots, a_{|\mathcal{A}|}\}$.
As shown in Algorithm \ref{alg:opt}, the search archive is initialized with the all-structure policy $\rho_{\mathrm{all}}$. An LLM proposer then generates executable selection policies under a prior direction of reflecting the selection mechanisms. These mechanisms capture how a policy infers query demand, estimates structure utility, constructs a subset, and balances robustness against evidence cost. Their complete descriptions of prior directions are provided in Appendix.

\MESA balances exploration and exploitation using an Upper Confidence Bound (UCB)-guided scheduling~\citep{auer2002finite,fialho2010analyzing}.
Each prior direction $a\in\mathcal{A}$ defines a way to generate or modify a selection policy and serves as one UCB arm. Let $n_a$ be the number of candidate policies evaluated under direction $a$. We compute
\begin{equation}
\mathrm{UCB}(a)
=
\overline{\widetilde{J}}_{\mathrm{val}}(a)
+
\beta
\sqrt{
\frac{\log(N+2)}{n_a+1}
},
\label{eq:mechanism-ucb}
\end{equation}
where $n_a$ is the number of evaluated candidates generated under
direction $a$, $N=\sum_a n_a$, and $\overline{\widetilde{J}}_{\mathrm{val}}(a)$ is their mean validation
objective. The first term
favors directions that have previously produced strong policies,
whereas the second encourages directions that have been evaluated less
frequently.

At each iteration, the LLM proposer receives the policy archive, the
UCB statistics of all directions, and the recurring errors of the
current policy. It then generates three executable candidates: an
exploitation candidate based on a previously effective direction, an
exploration candidate based on an under-explored direction, and a
failure-repair candidate targeting a recurring error. Thus,
Eq.~(\ref{eq:mechanism-ucb}) guides the choice of search directions,
whereas the LLM translates each direction into a concrete policy
modification. Each candidate is evaluated on the validation set, and
its result updates the corresponding direction's $\overline{\widetilde{J}}_{\mathrm{val}}(a)$ and
$n_a$. Further details are provided in Appendix.
After the search budget is exhausted, \MESA returns the policy with the highest objective score on the validation set in the archive (Algorithm 1, line 12). After optimization, the selected policy $\rho^*$ is frozen and tested on the held-out test set.
\begin{algorithm}[t]
\caption{Prior-Guided Harness Optimization}
\label{alg:opt}
\begin{algorithmic}[1]
\Require training set $\mathcal{D}_{\mathrm{tr}}$, validation set
$\mathcal{D}_{\mathrm{val}}$, proposer $\Pi$, prior directions
$\mathcal{A}$, budget $T$
\State $\mathcal{P}\gets
\{(\rho_{\mathrm{all}},
\textsc{Evaluate}(\rho_{\mathrm{all}},\mathcal{D}_{\mathrm{val}}))\}$
\For{$t=1,\ldots,T$}
    \State $\mathcal{C}_t\gets
    \textsc{Propose}_{\mathrm{UCB}}(\Pi,\mathcal{P},\mathcal{A})$
    \For{$\rho\in\mathcal{C}_t$}
        \If{\textsc{Valid}$(\rho)$}
            \State $\rho\gets
            \textsc{Fit}(\rho,\mathcal{D}_{\mathrm{tr}})$
            \State $E_\rho\gets
            \textsc{Evaluate}(\rho,\mathcal{D}_{\mathrm{val}})$
            \State $\mathcal{P}\gets\mathcal{P}\cup\{(\rho,E_\rho)\}$
        \EndIf
    \EndFor
\EndFor
\State \Return
$\displaystyle
\rho^*=
\arg\max_{(\rho,E_\rho)\in\mathcal{P}}
\widetilde{J}_{\mathrm{val}}(\rho)$
\end{algorithmic}
\end{algorithm}

\section{Experiment}
\label{sec:experiments}

\subsection{Experimental Setup}
\label{sec:experimental_setup}

\noindent\textbf{Datasets.}
\textbf{AMA-Bench} \cite{zhao2026ama} contains long agent--environment trajectories in which later questions require evidence produced during earlier reasoning, actions, observations, or tool calls. We use its real-world subset, consisting of 208 episodes with 12 expert-curated question--answer pairs per episode, for a total of 2,496 questions. The episodes cover six agentic domains: web task execution, open-world tool question answering, text-to-SQL, software engineering, gaming, and embodied AI. The questions additionally span four memory capabilities: recall, causal inference, state updating, and state abstraction. We further evaluate on \textbf{LoCoMo} \cite{maharana2024locomo} to test whether the same framework extends beyond agent trajectories to long-term conversational memory. LoCoMo contains 10 multi-session conversations and 1,986 question--answer pairs. Following the standard protocol, we exclude 446 adversarial questions and evaluate the remaining 1,540 questions: 841 single-hop, 282 multi-hop, 321 temporal, and 96 open-domain questions.\\
\noindent\textbf{Baselines.}
We compare against four families of baselines. \emph{Long-context} directly provides the trajectory to the answer model. \emph{BM25} and \emph{Qwen3-Emb-4B} represent sparse and dense item-level retrieval, respectively. We also include representative long-term memory systems: MemGPT, HippoRAG2, Mem0, MemoRAG, A-Mem, EMem, the multi-structure memory method Hindsight and the benchmark-specific state-of-the-art baseline AMA-Agent. 
All methods within an AMA-Bench backbone block use the same answer model and are evaluated with the same prompt. For methods that were not originally designed for these datasets, we preserve their core memory and retrieval mechanisms while adapting the prompts to the task-specific memory construction, input and output formats. Detailed descriptions of the baselines are provided in the Appendix.\\
\noindent\textbf{Metrics.}
Following the standard evaluation protocols of each benchmark, we use LLM-judged accuracy for AMA-Bench and F1 score for LoCoMo. The judge LLM backbone is the same as the answer model. All predictions are evaluated using a fixed judging protocol shared across methods. The evaluation prompt is provided in Appendix.
\\
\noindent \textbf{Implementation Details.}
We split AMA-Bench at the episode-level into training, validation, and test splits with the ratio 2:2:6. The split is drawn at random while ensuring full coverage of the entire dataset (all six domains, twelve tasks, and four QA types). Because \MESA requires training to evolve its selector, we repeat this random partitioning five times and report the mean over the five splits, which reduces sensitivity to any single train/validation/test assignment. For LoCoMo, we use a conversation-disjoint 2:2:6 split. All answer models use greedy decoding. We use the same LLM backbone for each meta-harness optimization; we run 30 iterations. All experiments are run on NVIDIA A100 GPUs. 
Memory schema formulations, prompts, and exact hyperparameters $\lambda$ and $\beta$ are in Appendix. 

\subsection{Main Results}

\begin{table*}[t]
\centering
\setlength{\tabcolsep}{4pt}
\begin{tabular}{lccccc}
\toprule
Method
& Recall
& Causal Inference
& State Updating
& State Abstraction
& \textbf{Overall} \\
\midrule

\multicolumn{6}{l}{
\textit{\textbf{LLM Backbone: Qwen3-32B} \cite{yang2025qwen3}}
} \\

Long-context
& $60.2_{\pm 1.2}$
& $51.9_{\pm 2.1}$
& $\underline{52.9}_{\pm 1.1}$
& $37.6_{\pm 1.5}$
& $52.6_{\pm 1.0}$ \\

BM25
& $37.3_{\pm 1.2}$
& $48.9_{\pm 1.1}$
& $36.4_{\pm 1.0}$
& $29.0_{\pm 1.8}$
& $38.5_{\pm 0.6}$ \\

Qwen3-Emb-4B \cite{zhang2025qwen3}
& $47.9_{\pm 0.6}$
& $50.7_{\pm 2.2}$
& $44.3_{\pm 2.4}$
& $\underline{41.8}_{\pm 2.0}$
& $46.6_{\pm 0.4}$ \\

MemGPT \cite{packer2023memgpt}
& $32.8_{\pm 2.0}$
& $37.9_{\pm 1.8}$
& $35.2_{\pm 1.4}$
& $28.4_{\pm 2.1}$
& $33.9_{\pm 0.5}$ \\

HippoRAG2 \cite{gutierrez2024hipporag}
& $49.7_{\pm 1.7}$
& $55.5_{\pm 1.0}$
& $42.3_{\pm 2.4}$
& $34.0_{\pm 2.7}$
& $46.6_{\pm 1.1}$ \\

Mem0 \cite{chhikara2025mem0}
& $30.1_{\pm 1.0}$
& $36.6_{\pm 1.2}$
& $28.3_{\pm 0.8}$
& $29.4_{\pm 2.2}$
& $31.1_{\pm 0.6}$ \\

MemoRAG \cite{qian2409memorag}
& $45.7_{\pm 0.9}$
& $48.3_{\pm 0.6}$
& $39.4_{\pm 1.2}$
& $33.6_{\pm 2.2}$
& $43.1_{\pm 0.7}$ \\

Hindsight \cite{latimer2025hindsight}
& $35.8_{\pm 1.0}$
& $45.4_{\pm 1.0}$
& $31.0_{\pm 1.3}$
& $30.7_{\pm 2.6}$
& $36.0_{\pm 0.5}$
\\

A-Mem \cite{xu2026mem}
& $47.6_{\pm 1.4}$
& $47.3_{\pm 1.1}$
& $41.0_{\pm 0.7}$
& $31.9_{\pm 1.8}$
& $43.2_{\pm 0.7}$ \\

EMem \cite{wang2026mem}
& $48.6_{\pm 1.2}$
& $42.8_{\pm 1.6}$
& $39.8_{\pm 1.2}$
& $28.2_{\pm 0.9}$
& $41.5_{\pm 0.6}$ \\

AMA-Agent \cite{zhao2026ama}
& $\underline{62.0}_{\pm 1.1}$
& $\mathbf{65.0}_{\pm 1.5}$
& $52.6_{\pm 2.2}$
& $39.8_{\pm 1.8}$
& $\underline{56.6}_{\pm 0.8}$ \\

\midrule



\textbf{MESA (ours)}
& $\mathbf{69.2}_{\pm 1.8}$
& $\underline{63.6}_{\pm 1.8}$
& $\mathbf{65.8}_{\pm 1.9}$
& $\mathbf{57.8}_{\pm 2.2}$
& $\mathbf{65.1}_{\pm 0.8}$ \\

\midrule

\multicolumn{6}{l}{
\textit{\textbf{LLM Backbone: Gemma-4-31B }\cite{team2026gemma}}
} \\

Long-context
& $63.0_{\pm 2.4}$
& $\underline{64.3}_{\pm 0.7}$
& $\underline{70.8}_{\pm 1.5}$
& $46.0_{\pm 2.3}$
& $62.5_{\pm 1.3}$ \\

BM25
& $24.8_{\pm 1.5}$
& $39.1_{\pm 1.6}$
& $26.7_{\pm 0.6}$
& $22.3_{\pm 1.3}$
& $28.3_{\pm 0.9}$ \\

Qwen3-Emb-4B \cite{zhang2025qwen3}
& $46.6_{\pm 0.8}$
& $52.5_{\pm 0.8}$
& $40.4_{\pm 1.8}$
& $33.1_{\pm 1.3}$
& $44.2_{\pm 0.7}$ \\

MemGPT \cite{packer2023memgpt}
& $36.1_{\pm 0.6}$
& $36.3_{\pm 3.3}$
& $42.2_{\pm 1.8}$
& $26.8_{\pm 2.1}$
& $36.2_{\pm 1.4}$ \\

HippoRAG2 \cite{gutierrez2024hipporag}
& $51.5_{\pm 1.6}$
& $45.1_{\pm 1.0}$
& $52.2_{\pm 4.0}$
& $39.7_{\pm 1.9}$
& $48.2_{\pm 1.3}$ \\

Mem0 \cite{chhikara2025mem0}
& $24.6_{\pm 0.8}$
& $31.8_{\pm 1.0}$
& $22.1_{\pm 1.2}$
& $24.8_{\pm 1.5}$
& $25.7_{\pm 0.4}$ \\

MemoRAG \cite{qian2409memorag}
& $49.0_{\pm 0.6}$
& $53.5_{\pm 1.6}$
& $44.1_{\pm 2.0}$
& $33.9_{\pm 1.4}$
& $46.3_{\pm 0.8}$ \\

Hindsight \cite{latimer2025hindsight}
& $39.5_{\pm 1.7}$ 
& $52.6_{\pm 1.1}$ 
& $36.8_{\pm 1.8}$ 
& $40.2_{\pm 1.4}$ 
& $42.1_{\pm 1.2}$ 
\\

A-Mem \cite{xu2026mem}
& $52.6_{\pm 1.3}$
& $54.0_{\pm 0.8}$
& $51.9_{\pm 2.1}$
& $35.7_{\pm 1.6}$
& $50.0_{\pm 0.9}$ \\

EMem \cite{wang2026mem}
& $61.5_{\pm 2.6}$
& $54.2_{\pm 1.5}$
& $63.2_{\pm 1.2}$
& $48.1_{\pm 2.5}$
& $58.0_{\pm 1.4}$ \\
AMA-Agent \cite{zhao2026ama}
& $\underline{66.1}_{\pm 1.0}$
& $62.2_{\pm 1.9}$
& $66.1_{\pm 2.2}$
& $\underline{53.3}_{\pm 1.5}$
& $\underline{63.0}_{\pm 1.0}$ \\
\midrule


\textbf{MESA (ours)}
& $\mathbf{69.9}_{\pm 2.2}$ & $\mathbf{68.8}_{\pm 1.1}$ & $\mathbf{73.9}_{\pm 2.3}$ & $\mathbf{62.0}_{\pm 1.9}$  &$\mathbf{69.4}_{\pm 1.1}$   \\

\bottomrule
\end{tabular}

\caption{
Main results on AMA-Bench using Qwen3-32B and Gemma-4-31B
as answer-model backbones. Results are LLM-as-judge accuracies
(\%). Baseline results are reported as
mean$_{\pm\mathrm{std}}$ over five splits. Best results within each backbone
are shown in bold, and second-best results are
underlined.
}
\label{tab:main}
\end{table*}

\begin{table}[t]
\centering
\small
\setlength{\tabcolsep}{8pt}
\begin{tabular}{lc}
\toprule
\textbf{Method} & \textbf{F1} \\
\midrule
Long-context
& $44.7$ \\
BM25
& $27.4$ \\
Qwen3-Emb-4B \cite{zhang2025qwen3}
& $36.7$ \\
HippoRAG2 \cite{gutierrez2024hipporag}
& $36.6$ \\
MemGPT \cite{packer2023memgpt}
& $46.8$ \\
Mem0 \cite{chhikara2025mem0}
& $37.2$ \\
A-Mem \cite{xu2026mem}
& $36.2$ \\
MemoRAG \cite{qian2409memorag}
& $25.4$ \\
Hindsight \cite{latimer2025hindsight}
& $46.3$ \\
EMem \cite{wang2026mem}
& $\underline{47.2}$ \\
AMA-Agent \cite{zhao2026ama}
& $32.6$ \\
\midrule
\textbf{MESA (ours)}
& $\mathbf{49.0}$ \\
\bottomrule
\end{tabular}
\caption{
Token-F1 score (\%) of \MESA on the long-conversation
dataset LoCoMo with Qwen3-32B.
}
\label{tab:locomo}
\end{table}


\noindent\textbf{\MESA improves agentic long-horizon memory.}
Table~\ref{tab:main} presents the main AMA-Bench results. With Qwen3-32B, \MESA reaches 65.1\% overall accuracy, outperforming the strongest prior memory baseline, AMA-Agent, by 8.5 points and the long-context reader by 12.5 points. \MESA is best on three of the four memory capabilities. On Causal Inference, it obtains 63.6\%, 1.4\% lower than AMA-Agent. The category-level gains therefore do not arise from one dominant question type. With Gemma-4-31B, \MESA obtains an overall 6.4\% higher accuracy compared with the strongest baseline. These results show that the benefits of structure-level selection are consistent across answer-model backbones.
\\
\noindent\textbf{Gains span diverse agentic domains.}
The domain breakdown in Figure~\ref{fig:ama_domain_radar} (full table in Appendix) shows that \MESA is the best method on four of the six domains with both model backbones. It improves over the strongest baseline by 4.1\% on Web, 8.9\% on Text2SQL, 15.5\% on Software, and 8.5\% on Embodied AI. On Open-World QA and Gaming, \MESA is second by only 0.7\% in each case. The particularly large Software and Text2SQL gains are consistent with the need to combine global state with exact identifiers, code locations, and temporally localized events.
\\
\noindent\textbf{The framework extends to conversational memory.}
Table~\ref{tab:locomo} applies the same framework and optimization procedure to LoCoMo. \MESA obtains an overall F1 of 49.0\%, improving over the strongest baseline that specialized for this dataset \cite{wang2026mem} by 1.8\%. \yaoqi{The gain seems small. How to explain this?} 

\subsection{Ablation Studies of Selection Strategy}
\begin{table}[t]
\centering
\setlength{\tabcolsep}{4pt}
\begin{tabular}{lccc}
\toprule
\textbf{Selection strategy}
& \textbf{\#Str.}
& \textbf{Evidence tok./q}
& \textbf{Accuracy} \\
\midrule
Summary& 1.0 & 5.7k & $57.7_{\pm 1.3}$ \\
Temporal KV& 1.0 & 2.8k & $51.4_{\pm 0.8}$ \\
Knowledge graph& 1.0 & 0.8k & $32.7_{\pm 1.2}$ \\
Vector DB& 1.0 & 3.5k & $46.6_{\pm 0.4}$ \\
Raw context& 1.0 & 10.5k & $48.2_{\pm 0.5}$ \\
\midrule
Random& 2.5 & 11.4k & $56.7_{\pm 1.2}$ \\
LLM zero-shot& 2.2 & 11.2k & $61.4_{\pm 0.8}$ \\
All& 5.0 & 18.7k & $63.7_{\pm 1.0}$ \\
\midrule
Route-to-one& 1.0 & 6.8k & $57.0_{\pm 3.2}$ \\
w/o priors& 2.8 & 11.5k & $63.1_{\pm 2.3}$ \\
w/o UCB& 3.0 & 12.4k & $64.3_{\pm 2.2}$ \\
\textbf{MESA (ours)}& 2.8& 11.0k & $\mathbf{65.1}_{\pm 0.8}$ \\
\bottomrule
\end{tabular}
\caption{
Ablation of selection on AMA-Bench with Qwen3-32B.
\#Str. denote the average number of structures, and Evidence tok./q means evidence tokens per query. The first block shows the performance of single structures, the second block shows the ablation of selection without learning, the third block shows the ablation of learning designs. \yaoqi{I think swapping the "Zero-shot" and "Route-to-one" rows would better align the table with the explanation text.}
}
\label{tab:ablation}
\end{table}

\begin{figure}[!t]
    \centering
    \includegraphics[width=\columnwidth]
    {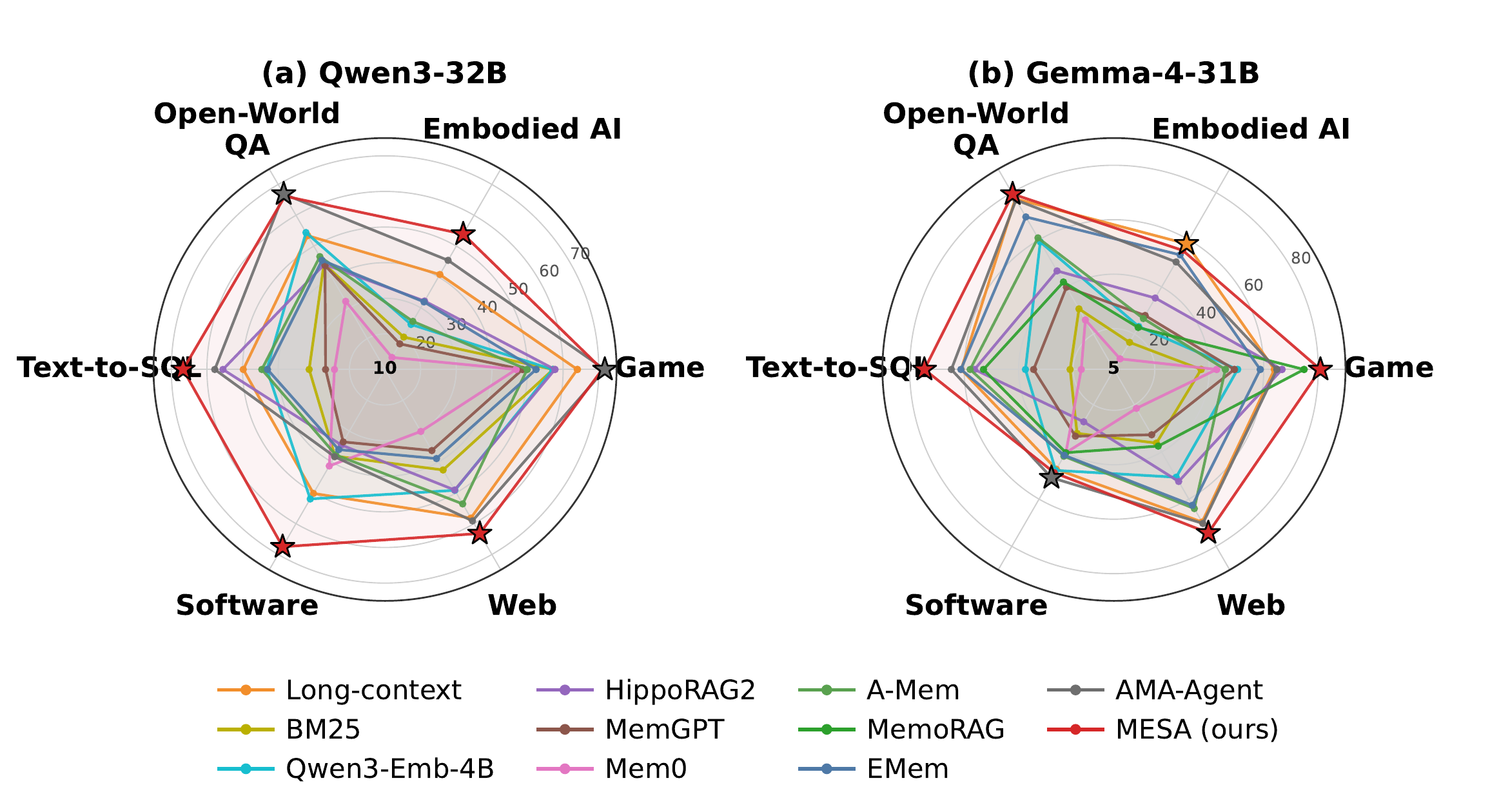}
    \caption{Domain-wise accuracy (\%) on AMA-Bench with Qwen3-32B and Gemma4-31B.}
    \label{fig:ama_domain_radar}
\end{figure}

\begin{figure}[!t]
    \centering
    \includegraphics[width=\columnwidth]
    {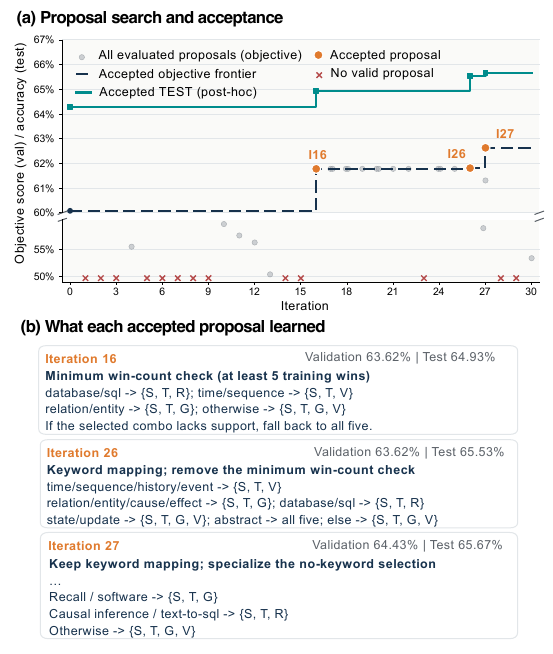}
    \caption{Example of \MESA's selection policy optimization on AMA-Bench. (a) Training-objective values for evaluated proposals, accepted archive updates, and the best-so-far policy. The accepted test accuracy is post-hoc accuracy. (b) Representative accepted proposals. }
\label{fig:optimization_trace}
\end{figure}
\noindent\textbf{Performance of individual structures.}
The first block of Table~\ref{tab:ablation} shows that no single
memory structure is sufficient across AMA-Bench. Summary is the
strongest individual structure, reaching 57.7\% accuracy with
5.7k evidence tokens, while the other structures achieve
32.7--51.4\%. Although the summary backend is already strong,
\MESA treats it as only one candidate and exceeds it by
7.4\% through selective combination with
complementary structures. Thus, the improvement cannot be
attributed to the underlying summary structure alone.\\
\noindent\textbf{Selection without learning.}
The second block compares selection strategies that do not use
the proposed optimization procedure. Random subsets achieve
56.7\% accuracy with 11.4k evidence tokens, whereas LLM
zero-shot selection reaches 61.4\% with a similar budget of
11.2k tokens, demonstrating the benefit of query-conditioned
selection. Reading all five structures further improves accuracy
to 63.7\%, but requires 18.7k evidence tokens per query. Exhaustive
access is therefore substantially more expensive and still does
not provide the best performance.\\
\noindent\textbf{Ablation of learning designs.}
The third block isolates the components of the learned selection
policy. Restricting the policy to route each query to only one
structure reduces accuracy to 57.0\%, confirming that adaptively
combining multiple structures is essential. Removing prior
directions lowers accuracy from 65.1\% to 63.1\%, showing that
structured guidance focuses proposals on plausible routing
modifications and makes sparse answer-level feedback more
actionable. Removing UCB lowers accuracy to 64.3\% and increases
the standard deviation from 0.8 to 2.2, indicating that UCB helps
balance exploration and exploitation across proposal directions
and stabilizes the search. With both components, \MESA selects
2.8 structures on average and reaches 65.1\% accuracy using
11.0k evidence tokens. It outperforms the all-structure
configuration by 1.4\% while using 41.2\% fewer
tokens, and improves over the zero-shot selector by 3.7\%.

\subsection{What does policy evolution learn?}
Fig.~\ref{fig:optimization_trace} shows a representative optimization run\footnote{Post-hoc test accuracy were generated after the complete optimization, which were never exposed during policy selection.}. Accepted proposals progressively refine keyword-based routing into query-conditioned rules, while rejected or invalid candidates leave the archived policy unchanged. The final policy achieves 65.67\% test accuracy, while remaining explicit and interpretable. 

\yaoqi{Adding a subsection discussing token efficiency and comparing \MESA with other memory evolution methods, such as M-Star, would further strengthen the experimental evaluation.}

\section{Conclusion}
In this paper, we studied how a long-horizon agent should read its accumulated memory when no single organization serves every query. Framing this as structure-level selection, we showed that the useful subset of memory structures is intermediate and query-dependent, and introduced \MESA, a five-structure framework that learns a query-adaptive selector from answer-level feedback while leaving the builders, retrievers, and answer model fixed. On AMA-Bench, \MESA surpasses the strongest baseline by 8.5\% while reading over 40\% fewer evidence tokens than the all-structure union, and the framework also extends to conversational memory on LoCoMo. Future work will jointly optimize memory construction and selection and model cross-structure evidence dependencies.


\bigskip
\newpage

\bibliography{main}

\newpage
\appendix
\section{Memory Construction}
An agent episode is a triple $e_i=(d_i,\tau_i,\mathcal{Q}_i)$, where $d_i\in\mathcal{D}$ is a domain label and $\tau_i=(u_{i,1},\ldots,u_{i,L_i})$ is the interaction history. Each step $u_{i,\ell}=(a_{i,\ell},o_{i,\ell})$ contains an action and an observation. The episode is associated with grounded QA instances $\mathcal{Q}_i=\{(q_{i,j},t_{i,j},y_{i,j}^{\star})\}_{j=1}^{m_i}$, and $y_{i,j}^{\star}$ denote the question, question type, and reference answer, respectively.

For each interaction history, \MESA constructs a fixed memory library

$$
\mathcal{M}_i=\left\{M_i^{(k)}=B_k(\tau_i)\right\}_{k=1}^{K},
$$
where $B_k$ and $M_i^{(k)}$ denote the fixed builder and stored representation of structure $k$. For question $q_{i,j}$, a selector with policy $\rho$ predicts a non-empty binary selection vector
$$
\mathbf{z}_{i,j}=S_\rho(q_{i,j},c_{i,j})$$
$$
\mathbf{z}_{i,j}\in\mathcal{Z}=
\left\{\mathbf{z}\in\{0,1\}^{K}:\lVert\mathbf{z}\rVert_0\geq1\right\},
$$
where $c_{i,j}$ contains compact previews of the evidence available from each structure. Each selected structure is accessed through its fixed mechanism $\Gamma_k$, and the resulting evidence is concatenated for the frozen answer model $F$:

$$
E_{i,j}(\mathbf{z}_{i,j})=
\operatorname{Compose}\!\left(
\left\{\Gamma_k(q_{i,j},M_i^{(k)}):z_{i,j}^{(k)}=1\right\}
\right),$$
$$
\hat{y}_{i,j}=F(q_{i,j},E_{i,j}).
$$
\MESA optimizes only $\rho$. The builders $B_k$, access mechanisms $\Gamma_k$, composer, and answer model remain fixed.

\subsection{Multi-view Structural Memory Construction}
We follow the notation in Eq. (1) of the main paper. For each interaction history $\tau_i$, \MESA constructs a fixed library $\mathcal{M}_i=\{M_i^{(k)}=B_k(\tau_i)\}_{k=1}^{K}$, where $B_k$ is the builder and $\Gamma_k$ is the corresponding access mechanism. The five structures are indexed by $k\in\{S,T,G,V,R\}$ for summary, temporal store, knowledge graph, vector database, and raw episodic trace, respectively. All structures are constructed offline and remain fixed while \MESA learns only the selector $S_\rho$.
\subsubsection{Summary (S) Memory}
Following AMA-Agent (Zhao et al., 2026), the summary builder $B_S$ compresses the interaction history $\tau_i=(u_{i,1},\ldots,u_{i,L_i})$ into an episode-level state memory, where each step $u_{i,j}=(a_{i,j},o_{i,j})$ contains an action and an observation. The summary preserves salient state changes, executed operations, results, errors, and subgoal transitions together with their temporal anchors, while removing repeated or task-irrelevant observations.

For a long history, $B_S$ partitions $\tau_i$ into $C_i$ consecutive, turn-aligned segments, summarizes them independently, and concatenates the partial summaries in temporal order:
$$
M_i^{(S)}=
\operatorname{Concat}_{c=1}^{C_i}
\left[f_{\mathrm{sum}}\!\left(\tau_i^{(c)}\right)\right],
$$
where a single segment is used when the history fits within the construction window. The resulting $M_i^{(S)}=B_S(\tau_i)$ provides broad episode context and remains fixed across queries. The prompt of constructing the summary memory is shown in \ref{prompt:summary}.

\subsubsection{Temporal Store (T) Memory}
The temporal builder $B_T$ preserves the original action--observation sequence as a turn-indexed store:
$$
M_i^{(T)}=B_T(\tau_i)=
\{(\ell,a_{i,\ell},o_{i,\ell})\}_{\ell=1}^{L_i}.
$$
The access mechanism returns a query-specific subset rather than the complete store:
$$
E_{i,j}^{(T)}=\Gamma_T(q_{i,j},M_i^{(T)})=
\{(\ell,a_{i,\ell},o_{i,\ell}):\ell\in\mathcal{I}_{i,j}\}.
$$
When the selector activates the temporal structure, $\Gamma_T$ first constructs $\mathcal{I}_{i,j}$ from turns explicitly referenced by the question, including individual steps, ranges, lists, and prefixes such as ``until step $\ell$,'' together with supplementary turns requested by AMA-Agent's sufficiency checker. Duplicate turns are removed and the retained action--observation records are ordered by step index. If no evidence can be localized through these temporal references, $\Gamma_T$ falls back to lexical matching over the temporal keys and returns up to five highest-ranked turn identifiers,
$$
E_{i,j}^{(T)}=\{\operatorname{key}(\ell):\ell\in
\operatorname{TopK}_{5}^{\mathrm{lex}}(q_{i,j})\}.
$$

\subsubsection{Graph (G) Memory }
The graph builder $B_G$ reuses the Qwen3-32B OpenIE artifacts produced by HippoRAG2 to form an episode-level relational graph. Each extracted edge records a subject, relation, object, and its associated trajectory step:
$$
M_i^{(G)}=B_G(\tau_i)=
\left(\mathcal{V}_i,\mathcal{E}_i\right),\qquad
\mathcal{E}_i=\{(h_\ell,r_\ell,t_\ell,j_\ell)\}_{\ell=1}^{N_i}.
$$
The step anchor $\ell_n$ is inferred from the source passage and preserves a link from each relation to the original execution history. At query time, \MESA uses a lightweight access mechanism compared to the original HippoRAG2. $\Gamma_G$ links query mentions to graph entities and scores relations using lexical and entity overlap. If the query contains a exact step or turn reference $r$, a relation anchored at $\ell_n$ receives an additional proximity score $1.25/(1+|\ell_n-r|)$; relations within one step of $r$ are also treated as step-local seeds. The matched entities are then expanded through one- and two-hop neighborhoods. The access mechanism returns up to $k_G=8$ highest-scoring focal relations, each augmented with up to four highest-scoring neighboring relations. This structure provides cross-step entity and causal connections that are difficult to recover from a flat sequence alone.

\subsubsection{Vector Database (V) Memory}
The summary structure is deliberately compressed and may omit a local detail that becomes relevant only after a query is observed. The vector builder $B_V$ therefore retains a dense, step-level view of the same history. Each turn is serialized with its temporal anchor, action, and observation:
$$
x_{i,j}=\operatorname{Serialize}(j,a_{i,j},o_{i,j}).
$$

Let $E_{\mathrm{emb}}$ denote the fixed embedding encoder. We normalize each vector and store it together with the source text and turn index:
$$
\mathbf{v}_{i,j}=
\frac{E_{\mathrm{emb}}(x_{i,j})}
{\lVert E_{\mathrm{emb}}(x_{i,j})\rVert_2},
$$
$$
M_i^{(V)}=B_V(\tau_i)=
\{(j,x_{i,j},\mathbf{v}_{i,j})\}_{j=1}^{L_i}.
$$
We instantiate $E_{\mathrm{emb}}$ with Qwen3-Embedding-4B. At query time, the fixed access mechanism $\Gamma_V$ encodes $q_i$ with the same encoder,
$$
\mathbf{v}_{q_i}=\frac{E_{\mathrm{emb}}(q_i)}
{\lVert E_{\mathrm{emb}}(q_i)\rVert_2},
$$
and returns the $k_V$ turns with the largest inner products $\mathbf{v}_{q_i}^{\top}\mathbf{v}_{i,j}$. Because both vectors are L2-normalized, this score equals cosine similarity. We use $k_V=5$. This representation preserves local evidence that is semantically related to the query, while the stored turn indices maintain the connection between each retrieved item and the original interaction history.

\subsubsection{Raw Episodic (R) Memory}
The raw episodic builder $B_R$ serializes the original actions and observations into overlapping, turn-aligned windows without semantic compression. For a trajectory of length $L_i$, the window size $w_i$ and overlap $d_i$ are
$$
w_i=\operatorname{clip}\!\left(\left\lceil\frac{L_i}{12}\right\rceil,3,16\right),
\qquad
d_i=\operatorname{round}(0.25w_i).
$$
Using stride $w_i-d_i$, the builder produces
$$
M_i^{(R)}=B_R(\tau_i)=
\{(s_c,e_c,x_{i,c})\}_{c=1}^{C_i},
$$
where $s_c$ and $e_c$ are the start and end positions in the trajectory sequence, and $x_{i,c}$ is the corresponding serialized action--observation text. When the selector activates this structure, $\Gamma_R$ ranks all windows against $q_{i,j}$ with BM25 and returns up to $k_R=3$ positive-scoring windows. If no window receives a positive score, it returns the single highest-ranked window. This structure preserves exact surface-form evidence and broader local context, complementing the compressed summary and more selective structured views.

\section{Baselines and Implementation Details}
\subsection{Baseline Methods}
\noindent\textbf{Long-context.} We feed the full historical trajectory into the backbone model. For trajectories exceeding the maximum context window, we truncate them to preserve the most recent history. 

\noindent\textbf{BM25}.We use the official BM25 implementation and its original configuration. Each trajectory is divided into steps containing the step index, action, and observation. Then, each step is further divided into non-overlapping chunks of at most 800 characters. We construct an independent \texttt{BM25Okapi} index for each episode using \texttt{rank-bm25} v0.2.2, lowercase whitespace tokenization, and the default parameters $k_1=1.5$, $b=0.75$, and $\epsilon=0.25$. For each question, the top-5 chunks are concatenated for answer generation. 

\noindent\textbf{Qwen3-Emb-4B} \cite{zhang2025qwen3}.
This baseline represents memory as a flat vector index over fine-grained history units, without additional summarization, graph construction, or memory evolution. On AMA-Bench, the task description and individual action--observation steps are encoded using Qwen3-Embedding-4B into 2,560-dimensional vectors and stored in a FAISS inner-product index. All embeddings are L2-normalized, making inner product equivalent to cosine similarity. For each question, we encode the query with the same model, retrieve the top five trajectory steps, concatenate them in similarity order, and pass the resulting context to the backbone LLM for answer generation. We set the embedding batch size to 8 and the maximum embedding length to 512 tokens. On LoCoMo, the same flat vector-memory structure is used, but each timestamped dialogue turn forms one indexed unit and the retrieval depth is increased to $k=10$. The query is prefixed with a retrieval instruction, and the retrieved turns retain their timestamps, dialogue identifiers, speakers, and original text.

\noindent\textbf{HippoRAG2} \cite{gutierrez2024hipporag}.
HippoRAG2 organizes memory as a heterogeneous graph consisting of passage, entity, and relational-fact nodes. OpenIE first extracts entities and relation triples from each passage, after which edges connect facts to their supporting passages and shared entities, while additional similarity edges connect synonymous entities. During retrieval, the query is linked to the top-$k$ fact or entity nodes, and relevance is propagated through the graph using personalized PageRank; the resulting graph scores are combined with recognition-memory reranking and dense passage scores. We preserve the original graph construction and retrieval settings, including linking top-$k=5$, QA top-$k=5$, PageRank damping factor $0.5$, and passage-node weight $0.05$. On AMA-Bench, consecutive trajectory steps are packed into OpenIE documents of at most 24,000 characters, with a 300-character overlap when an individual step must be split. The backbone LLM performs OpenIE at temperature $0$ with a 1,024-token limit, and Contriever provides graph embeddings. We retrieve 30 passage candidates and provide the top five, capped at 60,000 characters, for answering. On LoCoMo, each dialogue turn forms one passage, BAAI/bge-m3 is used for embeddings, and 200 passage candidates are retrieved before the top five are passed to the answerer.

\noindent\textbf{MemGPT} \cite{packer2023memgpt}.
MemGPT is a hierarchical memory system that organizes memory as a two-tier hierarchical structure inspired by virtual memory systems: a small in-context core memory serves as working memory, while a vector-indexed archival store maintains the complete long-term history. The backbone LLM explicitly searches the archival store and temporarily loads retrieved passages into its active context for reasoning. On AMA-Bench, the task description is placed in core memory, while the complete trajectory is divided into non-overlapping passages of at most 1,500 characters and stored in archival memory with normalized Qwen3-Embedding-4B vectors. For each question, the backbone LLM can iteratively issue search queries. Each search retrieves the five most similar passages using cosine similarity and returns at most 8,000 characters to the core context. We manually set the maximum number of searches to five, the generation budget of each agent step to 512 tokens, and the temperature to $0$. If no valid answer is produced within the search budget, the backbone LLM is forced to answer from the accumulated retrieved context. On LoCoMo, we use the same two-level memory structure and retrieval parameters, but archival passages contain timestamped dialogue turns rather than trajectory steps, and each agent step is limited to 256 output tokens. Archival memory remains fixed during evaluation and is not updated across questions.

\noindent\textbf{Mem0} \cite{chhikara2025mem0}.
Mem0 represents memory as a user-scoped collection of atomic facts stored in a vector database. The backbone LLM extracts candidate facts from each input chunk and compares them with existing memories to determine whether each fact should be added, updated, merged, or discarded. Each retained fact is stored with its embedding and metadata in an on-disk Qdrant database. On AMA-Bench, we build one isolated memory store per episode and group trajectory steps into chunks of at most 10 turns or 40,000 characters. We set the fact-extraction budget to 4,096 tokens, the embedding-input limit to 20,000 tokens, and retrieve the top 10 facts for each question. The backbone LLM performs fact extraction and conflict resolution at temperature $0$, while Qwen3-Embedding-4B provides 2,560-dimensional embeddings. On LoCoMo, following the official Mem0 conversational setup, we maintain one user-specific store for each conversation participant. Dialogue messages are inserted in two-message batches with session timestamps as metadata, and the top 10 facts are retrieved from each participant's store. The retrieved facts are concatenated and passed to the backbone LLM for answer generation.

\noindent\textbf{A-Mem} \cite{xu2026mem}. A-Mem organizes memory as a dynamic graph-like structure of interconnected notes. Each trajectory step is represented as a memory node containing the original content together with an LLM-generated contextual description, keywords, and tags. Edges connect semantically related notes, allowing information from distant trajectory steps to be associated. When a new note is added, A-Mem performs memory evolution by deciding whether to create new links, strengthen existing relations, or update neighboring notes. During retrieval, the question is converted into keywords, which are used to retrieve the top-$k$ memory nodes together with their linked neighbors. We use the official robust A-Mem implementation with the backbone LLM for note construction and evolution, all-MiniLM-L6-v2 for node embeddings, and $k=10$. The maximum note and retrieval-context lengths are both 120,000 characters, and metadata generation is limited to 1,000 tokens with temperature $0$. For LoCoMo, the same memory structure and parameters are used, but each timestamped dialogue turn forms a memory node instead of each trajectory step.

\noindent\textbf{MemoRAG} \cite{qian2409memorag}.
MemoRAG uses a hybrid memory structure that combines a compressed global memory state with a dense retrieval index. The complete trajectory is first encoded by the MemoRAG memory model, \texttt{memorag-qwen2-7b-inst}, into a compact hidden-state memory using beacon-based context compression. In parallel, the trajectory is divided into 512-token chunks and indexed by BAAI/bge-m3. For each question, the memory model recalls potentially relevant text spans and generates surrogate queries, which are used to retrieve the top three chunks from the dense index. The retrieved evidence is then passed to the backbone LLM for answer generation. For each question, the memory model recalls relevant text spans and generates surrogate queries. Each query retrieves its top three 512-token chunks from the dense index, and the deduplicated union of the retrieved chunks is passed to the backbone LLM.

\noindent\textbf{Hindsight} \cite{latimer2025hindsight}.
Hindsight organizes knowledge in a memory bank containing four structured memory types: \textit{world facts} for objective knowledge, \textit{experience facts} for the agent's actions and interactions, \textit{observations} for automatically consolidated knowledge derived from multiple facts, and \textit{mental models} for higher-level summaries of recurring queries or concepts. During retention, the backbone LLM extracts facts, entities, relations, and temporal information from input documents and stores them in the corresponding memory structures. During recall, Hindsight searches memories through four parallel strategies: semantic similarity, BM25 keyword matching, graph traversal, and temporal filtering. Then it combines their results using reciprocal-rank fusion and cross-encoder reranking. On AMA-Bench, we construct one memory bank per episode, divide trajectories into documents of at most 4,000 characters, and truncate trajectories longer than 120,000 characters using a head--tail strategy. On LoCoMo, each timestamped conversation session is retained as one document in a conversation-specific bank. For both datasets, we use the \texttt{high} recall budget and an 8,192-token recall limit. In our configuration, automatic observation consolidation is disabled and no task-specific mental models are provided; therefore, evaluation primarily relies on the world-fact and experience-fact memory structures.

\noindent\textbf{E-Mem} \cite{wang2026mem}.
E-Mem organizes memory as a hierarchy of episodic blocks coordinated by a master--assistant architecture. Each block is managed by an assistant agent and preserves its original, uncompressed context together with an LLM-generated summary, while the master agent coordinates retrieval and aggregates evidence returned by the selected assistants. In our implementation, each block contains approximately 4,096 tokens, corresponding to $12.5\%$ of the backbone LLM's 32,768-token context window, and adjacent blocks have a $10\%$ overlap. For each question, a hybrid router scores blocks using summary-level semantic similarity, chunk-level semantic similarity, and BM25 matching with weights $0.3$, $0.4$, and $0.3$, respectively. The semantic text index uses chunks of size 512 with an overlap of 50, and Qwen3-Embedding-4B provides the embeddings. The router selects at most five memory blocks. Each selected assistant independently reconstructs relevant evidence from its local context, and the backbone LLM aggregates the returned evidence and generates the final answer. On AMA-Bench, trajectory steps are inserted sequentially into an episode-specific memory hierarchy, whereas on LoCoMo, timestamped dialogue turns are inserted into a conversation-specific hierarchy. All remaining memory-block and routing parameters are shared across the two benchmarks.

\noindent\textbf{AMA-Agent} \cite{zhao2026ama}.
AMA-Agent maintains a heterogeneous memory structure consisting of a compressed state memory, the original trajectory store, and a dense index over individual trajectory turns; it can additionally construct a causal graph, which is disabled in our experiments. The state memory summarizes the task, important entities, intermediate states, and critical events, while the raw trajectory and turn embeddings preserve fine-grained evidence that may be lost during compression. During retrieval, AMA-Agent first retrieves the top five turns using Qwen3-Embedding-4B and directly includes any turns explicitly referenced by the question. The backbone LLM then judges whether the retrieved evidence is sufficient. If it is not sufficient, it can request adjacent or ranged turns through structured retrieval, or generate and execute a Python search program for trajectory-wide counting and aggregation. We set the construction chunk size to 2,048, the session size to 8,192, the final state-memory budget to 1,000 tokens, the retrieval depth to $k=5$, and the temperature to $0$. On AMA-Bench, memory is constructed from action--observation trajectory steps and includes the task description. On LoCoMo, each timestamped dialogue turn is represented as a trajectory step with a fixed \texttt{dialogue} action, while the same memory structure and retrieval parameters are retained.

\subsection{Hyperparameter Selection and Budget}
\label{app:hyperparameter_selection}

\MESA introduces two optimization-specific scalar
hyperparameters: the evidence-cost coefficient $\lambda$ in
the regularized validation objective and the exploration
coefficient $\beta$ in the UCB scheduler. These values are
determined using development data only and are fixed before
evaluation on the test split. We use the same values across
all data splits, answer-model backbones, and benchmarks,
rather than retuning them for each experimental setting.
The memory builders, access mechanisms, and retrieval
depths are also fixed independently of selector optimization,
as described in Appendix~A.

\paragraph{Evidence-cost coefficient.}
We set $\lambda=10^{-6}$ by calibrating the token penalty
to the scale of the answer-level evaluation score. Since
$J_{\mathrm{val}}(\rho)\in[0,1]$ while
$C_{\mathrm{val}}(\rho)$ is measured in raw evidence tokens,
this value assigns a penalty of $0.01$ to an additional
$10{,}000$ evidence tokens per query:
\[
10^{-6}\times 10{,}000 = 0.01.
\]
Thus, an increase of approximately $10{,}000$ tokens must
provide at least one percentage point of validation-accuracy
improvement to be favored by the regularized objective.
This calibration keeps the cost term large enough to
discourage unnecessarily exhaustive retrieval without
allowing small token differences to dominate answer quality.

\paragraph{UCB exploration coefficient.}
We set $\beta=0.08$ so that the exploration bonus remains
comparable to the differences in regularized validation
scores observed among candidate policies. A substantially
smaller value makes scheduling nearly greedy and repeatedly
selects directions that perform well early in the search,
whereas an excessively large value over-prioritizes
under-evaluated directions regardless of their observed
utility. Because an unvisited direction is initialized with
the current frontier score, $\beta$ controls the additional
uncertainty bonus rather than assigning an artificially low
initial value to unexplored directions. The ablation with
$\beta=0$, reported as \emph{w/o UCB} in Table~3, further
shows that removing this exploration mechanism reduces
accuracy and increases variance across splits.

\paragraph{Optimization budget.}
We use a fixed budget of $T=30$ optimization iterations.
At each iteration, the proposer generates one exploitation,
one exploration, and one failure-repair candidate. The same
budget is used for \MESA and all optimization ablations, so
their differences cannot be attributed to additional proposal
or validation calls. The budget is treated as a computational
constraint rather than a parameter selected using test
performance. After the budget is exhausted, the policy with
the highest regularized validation objective in the archive is
frozen and evaluated once on the held-out test split.

\section{Prompts}
\subsection{Summary (S) Prompt}
We use the same prompt for constructing the text summary.
\begin{tcolorbox}[
    enhanced,
    breakable,
    title={Summary (S) Memory Construction Prompt},
    colback=blue!5,
    colframe=black!60,
    boxrule=0.6pt,
    arc=1mm,
    left=2mm,
    right=2mm,
    top=1.5mm,
    bottom=1.5mm,
    fonttitle=\bfseries,
]
\small
\raggedright
\label{prompt:summary}
COMPRESS\_PROMPT\_TEMPLATE = """You are presented with a section of agent trajectory (actions and observations). Compress it into a state memory that future readers can use to answer detailed questions about what happened.

\medskip
Task: \{task\}

\medskip
Trajectory Section:\\
\{trajectory\_text\}

\medskip
\{previous\_state\_text\}

\medskip
Identify KEY turns -- turns where state meaningfully changed, a command/query/code was executed, an important result/value/schema/error appeared, or the agent shifted sub-goal. For each key turn record an env\_state block with bullet keys that BEST FIT this turn. Common keys include: action, command, query, location, inventory, schema, sql, result, finding, error, change. Use whatever keys fit; do NOT force fields that do not apply, and do NOT dump unrelated context into a "change" field.

\medskip
CRITICAL -- copy these VERBATIM from the trajectory; never paraphrase or summarize:\\
- commands, queries, code (SQL, shell, Python, function/tool calls)\\
- identifiers: file paths, URLs, table/column/schema names, IDs, UI element ids and selectors, entity names\\
- numeric values, counts, computed results, dates, response codes\\
- error messages

\medskip
Also write a one-line Memory Summary describing the section's overall progress.

\medskip
Output everything after the marker below.

\medskip
\textbf{STATE\_MEMORY}

\medskip
memory\_summary: key\_turns=<turn ids>; key\_objective=<main objective>; key\_events=<critical events, commands, state changes, key results>

\medskip
turn\_id: <turn number>\\
env\_state:\\
- <key>: <value>\\
- ...

\medskip
turn\_id: <turn number>\\
env\_state:\\
- <key>: <value>

\medskip
"""

\end{tcolorbox}

\subsection{Graph (G) Memory Prompts}
We construct the knowledge-graph memory using the two-stage OpenIE
pipeline of HippoRAG2 \cite{gutierrez2025rag}. Given a packed trajectory passage, the first
LLM call extracts a list of named entities. The second call conditions
on both the original passage and the extracted entity list to generate
subject--relation--object triples. The two prompts are shown below.
We use Qwen3-32B with temperature $0$, disabled reasoning, and a
maximum output length of 1,024 tokens. Each packed trajectory passage
contains at most 24,000 characters.
\begin{tcolorbox}[
    title={Named Entity Extraction Prompt},
    colback=pink!10,          
    colframe=pink!60!black,   
    boxrule=0.6pt,
    arc=1mm,
    left=2mm,
    right=2mm,
    top=1.5mm,
    bottom=1.5mm,
    fonttitle=\bfseries,
]
\small
\noindent\textbf{[System]} \\
Your task is to extract named entities from the given paragraph. \\
Respond with a JSON list of entities.

\medskip
\noindent\textbf{[User: one-shot example]} \\
Radio City \\
Radio City is India's first private FM radio station and was started on 3 July 2001. \\
It plays Hindi, English and regional songs. \\
Radio City recently forayed into New Media in May 2008 with the launch of a music portal - PlanetRadiocity.com that offers music related news, videos, songs, and other music-related features.

\medskip
\noindent\textbf{[Assistant: one-shot output]} \\
\texttt{\{"named\_entities": ["Radio City", "India", "3 July 2001", "Hindi", "English", "May 2008", "PlanetRadiocity.com"]\}}

\medskip
\noindent\textbf{[User]} \\
\texttt{\$\{passage\}}
\end{tcolorbox}

\begin{tcolorbox}[
    enhanced,
    breakable,
    title={Triple Extraction Prompt},
    colback=pink!10,
    colframe=pink!60!black,
    boxrule=0.6pt,
    arc=1mm,
    left=2mm,
    right=2mm,
    top=1.5mm,
    bottom=1.5mm,
    fonttitle=\bfseries,
]
\small
\raggedright

\noindent\textbf{[System]} \\
Your task is to construct an RDF (Resource Description Framework)
graph from the given passages and named entity lists. \\
Respond with a JSON list of triples, with each triple representing
a relationship in the RDF graph.

\medskip
Pay attention to the following requirements: \\
- Each triple should contain at least one, but preferably two, of the
named entities in the list for each passage. \\
- Clearly resolve pronouns to their specific names to maintain clarity.

\medskip
\noindent\textbf{[User: one-shot example]} \\
Convert the paragraph into a JSON dict, it has a named entity list
and a triple list. \\
Paragraph: \\
\texttt{Radio City} \\
\texttt{Radio City is India's first private FM radio station and was} \\
\texttt{started on 3 July 2001.} \\
\texttt{It plays Hindi, English and regional songs.} \\
\texttt{Radio City recently forayed into New Media in May 2008 with} \\
\texttt{the launch of a music portal - PlanetRadiocity.com that offers} \\
\texttt{music related news, videos, songs, and other music-related features.}

\medskip
\texttt{\{"named\_entities":} \\
\texttt{["Radio City", "India", "3 July 2001", "Hindi",} \\
\texttt{"English", "May 2008", "PlanetRadiocity.com"]\}}

\medskip
\noindent\textbf{[Assistant: one-shot output]} \\
\texttt{\{"triples": [} \\
\texttt{["Radio City", "located in", "India"],} \\
\texttt{["Radio City", "is", "private FM radio station"],} \\
\texttt{["Radio City", "started on", "3 July 2001"],} \\
\texttt{["Radio City", "plays songs in", "Hindi"],} \\
\texttt{["Radio City", "plays songs in", "English"],} \\
\texttt{["Radio City", "forayed into", "New Media"],} \\
\texttt{["Radio City", "launched", "PlanetRadiocity.com"],} \\
\texttt{["PlanetRadiocity.com", "launched in", "May 2008"],} \\
\texttt{["PlanetRadiocity.com", "is", "music portal"],} \\
\texttt{["PlanetRadiocity.com", "offers", "news"],} \\
\texttt{["PlanetRadiocity.com", "offers", "videos"],} \\
\texttt{["PlanetRadiocity.com", "offers", "songs"]} \\
\texttt{]\}}

\medskip
\noindent\textbf{[User]} \\
Convert the paragraph into a JSON dict, it has a named entity list
and a triple list. \\
Paragraph: \\
\texttt{\$\{passage\}}

\medskip
\texttt{\$\{named\_entity\_json\}}

\end{tcolorbox}
The triple-extraction prompt produces only subject--relation--object
triples. It does not request trajectory-step indices. We associate each
triple with a step after extraction by matching its entities and relation
against the \texttt{Step N} blocks in the corresponding packed source
passage.

\subsection{Answer Generation Prompt}
\begin{tcolorbox}[
    enhanced,
    breakable,
    title={Answer Generation Prompt},
    colback=gray!3,
    colframe=black!60,
    boxrule=0.6pt,
    arc=1mm,
    left=2mm,
    right=2mm,
    top=1.5mm,
    bottom=1.5mm,
    fonttitle=\bfseries,
]
\small
\raggedright

\noindent\textbf{[User]} \\

\medskip
\noindent\textbf{\# Task} \\
\texttt{\$\{task\}}

\medskip
\noindent\textbf{\# State Memory (compressed)} \\
\texttt{\$\{B1\_evidence\}}

\medskip
\noindent\textbf{\# Temporal Turns} \\
\textbf{(referenced by step/turn or neighborhood } \\
\texttt{\$\{B2\_evidence\}}

\medskip
\noindent\textbf{\# Knowledge Graph} \\
\textbf{(HippoRAG triples + neighbors)} \\
\texttt{\$\{B3\_evidence\}}

\medskip
\noindent\textbf{\# Retrieved Turns} \\
\textbf{(embedding similarity)} \\
\texttt{\$\{B4\_evidence\}}

\medskip
\noindent\textbf{\# Chunked Context} \\
\textbf{(BM25 over trajectory)} \\
\texttt{\$\{B5\_evidence\}}

\medskip
\noindent\textbf{\#\# Questions} \\
Question: \texttt{\$\{question\}}

\medskip
\noindent\textbf{\#\# Instructions} \\
Provide a direct and concise answer.

\medskip
\noindent Answer: [your answer here]

\end{tcolorbox}
Only the memory structures selected are included in the
prompt. Empty or unselected sections are omitted. The complete prompt is sent
as a single user message without an additional system message.

\subsection{Judge Prompt}
We use the same LLM-as-a-judge prompt as \cite{zhao2026ama}, which is shown below.
\begin{tcolorbox}[
    title={Binary Correctness Judgement Prompt},
    colback=gray!3,
    colframe=black!60,
    boxrule=0.6pt,
    arc=1mm,
    left=2mm,
    right=2mm,
    top=1.5mm,
    bottom=1.5mm,
    fonttitle=\bfseries,
]
\small
\noindent\textbf{Prompt for Evaluating Answers for AMA-Bench} \\
You are an expert evaluator. You will be given a question, a reference answer, and a predicted answer. Your task is to determine if the predicted answer is correct based on:
\begin{enumerate}
    \setlength{\itemsep}{1pt}
    \setlength{\topsep}{2pt}
    \setlength{\leftmargin}{12pt}
    \item Factual correctness compared to the reference
    \item Completeness of the answer
    \item Relevance to the question
\end{enumerate}

\noindent\texttt{\{context\_str\}}

\medskip
\noindent\textbf{Question:} \texttt{\{question\}}

\medskip
\noindent\textbf{Reference Answer:} \texttt{\{golden\_answer\}}

\medskip
\noindent\textbf{Predicted Answer:} \texttt{\{predicted\_answer\}}

\medskip
\noindent Is the predicted answer correct? Respond with ONLY \texttt{yes} or \texttt{no}. Do not include any thinking process, explanation, or additional text.

\medskip
\noindent\textbf{Answer}
\end{tcolorbox}

\section{Prior Directions for Guided Optimization}
\label{app:prior}
The prior directions $\mathcal{A}=\{a_1,\ldots,a_M\}$ define
mechanism classes for proposing structure selectors; they do not encode
which memory subset is optimal for a particular question. Each direction
specifies an observable routing signal and a mechanism for mapping that
signal to a non-empty structure subset. Router parameters may be fitted
using training-split combination outcomes, while candidate acceptance and
frontier selection use only the validation objective. Held-out test
outcomes are excluded from the prior, proposer context, fitting procedure,
and optimization loop.

At inference time, the selector may access the question, task metadata, and
query-conditioned evidence previews. These previews are produced by the
fixed retrievers without access to reference answers, judge feedback, or
per-combination scores. Outcome-derived variables, including winning
combinations and cached evaluation scores, are available only as training
labels during learning and are never exposed to the selector.

The mechanism taxonomy is fixed before each optimization run. It restricts
the search to structure selection while leaving memory construction,
retrieval, composition, and answer generation unchanged. Additional
mechanism classes may be registered as new UCB arms before a run, but the
arm set is not modified using validation or test outcomes during that run.


\begin{axisbox}{Axis A --- Demand inference: \emph{what the query requires}}
\small
\textbf{(1) capability prior}, \textbf{(2) domain--task prior},
\textbf{(3) lexical demand inference}, and
\textbf{(4) query decomposition}: infer memory demand from the
question type, task metadata, lexical and identifier cues, or
decomposed sub-questions.
\end{axisbox}

\begin{axisbox}{Axis B --- Backend value estimation:
\emph{how useful each structure is}}
\small
\textbf{(5) marginal value}, \textbf{(6) conditional value},
\textbf{(7) hierarchical shrinkage}, and
\textbf{(8) learned gating}: estimate the utility of each structure
individually or conditioned on the current subset, while shrinking
low-support estimates toward coarser statistics.
\end{axisbox}

\begin{axisbox}{Axis C --- Subset construction:
\emph{how to form the selected set}}
\small
\textbf{(9) independent gating}, \textbf{(10) cardinality-first
selection}, \textbf{(11) forward selection}, and
\textbf{(12) complementarity selection}: construct a non-empty subset
through per-structure decisions, predicted subset size, incremental
addition, or complementary evidence coverage.
\end{axisbox}

\begin{axisbox}{Axis D --- Evidence-aware selection:
\emph{what the retrieved previews contain}}
\small
\textbf{(13) evidence availability}, \textbf{(14) evidence relevance},
\textbf{(15) evidence coverage}, and
\textbf{(16) evidence redundancy}: use query-conditioned previews to
remove empty or irrelevant structures, cover distinct information
needs, and suppress redundant evidence.
\end{axisbox}

\begin{axisbox}{Axis E --- Robustness and cost shaping:
\emph{how to control risk and context cost}}
\small
\textbf{(17) confidence fallback}, \textbf{(18) disagreement policy},
\textbf{(19) parsimony control}, and
\textbf{(20) support-aware routing}: fall back under uncertainty,
resolve conflicting routing signals, penalize unnecessary structures,
and back off from estimates supported by too few training examples.
\end{axisbox}

\section{Additional Result}
\subsection{Domain-wise Comparison on AMA-Bench}
We show the full result of Fig.~\ref{fig:ama_domain_radar} in Table~\ref{tab:main-ama-domain}.
\begin{table*}[t]
\centering
\setlength{\tabcolsep}{2.0pt}
\renewcommand{\arraystretch}{1.05}

\begin{tabular*}{\textwidth}{
@{\extracolsep{\fill}}lccccccc@{}
}
\toprule
\textbf{Method}
& \textbf{Web}
& \shortstack{\textbf{Open-world}\\\textbf{QA}}
& \textbf{Text2SQL}
& \textbf{Software}
& \textbf{Gaming}
& \shortstack{\textbf{Embodied}\\\textbf{AI}}
& \textbf{Overall} \\
\midrule

\multicolumn{8}{l}{
\textbf{Qwen3-32B \cite{yang2025qwen3}}
} \\
\addlinespace[1pt]

Long-context
& $58.2_{\pm 2.9}$ 
& $53.4_{\pm 3.2}$
& $49.8_{\pm 1.3}$
& $50.2_{\pm 2.2}$
& $64.0_{\pm 2.0}$
& $40.8_{\pm 4.0}$
& $52.6_{\pm 1.0}$ \\

BM25
& $42.6_{\pm 0.8}$
& $44.3_{\pm 2.7}$
& $31.3_{\pm 2.1}$
& $38.1_{\pm 0.2}$
& $57.2_{\pm 3.5}$
& $20.5_{\pm 0.9}$
& $38.5_{\pm 0.6}$ \\

Qwen3-Emb-4B \cite{zhang2025qwen3}
& $49.2_{\pm 0.9}$
& $54.4_{\pm 1.7}$
& $43.2_{\pm 1.2}$
& $\underline{52.0}_{\pm 1.5}$
& $57.1_{\pm 2.9}$
& $24.6_{\pm 2.1}$
& $46.6_{\pm 0.4}$ \\

HippoRAG2 \cite{gutierrez2024hipporag}
& $49.1_{\pm 2.1}$
& $44.1_{\pm 1.7}$
& $55.5_{\pm 0.7}$
& $34.5_{\pm 2.7}$
& $57.7_{\pm 3.0}$
& $32.2_{\pm 2.5}$
& $46.6_{\pm 1.1}$ \\

MemGPT \cite{packer2023memgpt}
& $36.3_{\pm 1.2}$
& $43.7_{\pm 2.0}$
& $26.7_{\pm 1.7}$
& $33.5_{\pm 1.8}$
& $48.6_{\pm 1.8}$
& $18.3_{\pm 0.5}$
& $33.9_{\pm 0.5}$ \\

Mem0 \cite{chhikara2025mem0}
& $30.1_{\pm 2.4}$
& $32.1_{\pm 1.2}$
& $24.2_{\pm 1.2}$
& $41.3_{\pm 2.3}$
& $46.8_{\pm 1.2}$
& $13.9_{\pm 1.4}$
& $31.1_{\pm 0.6}$ \\

Hindsight \cite{latimer2025hindsight}
&$39.5_{\pm 2.2}$
&$38.6_{\pm 1.7}$
&$29.8_{\pm 1.0}$
&$43.8_{\pm 1.9}$
&$50.7_{\pm 2.2}$
&$16.6_{\pm 2.0}$
&$36.0_{\pm 0.5}$ \\

A-Mem \cite{xu2026mem}
& $53.6_{\pm 2.5}$
& $46.6_{\pm 1.4}$
& $44.6_{\pm 1.2}$
& $37.7_{\pm 2.1}$
& $49.9_{\pm 2.1}$
& $25.6_{\pm 1.8}$
& $43.2_{\pm 0.7}$ \\

MemoRAG \cite{qian2409memorag}
& $38.7_{\pm 0.6}$ 
& $45.5_{\pm 1.7}$ 
& $47.8_{\pm 1.5}$ 
& $36.2_{\pm 2.1}$ 
& $63.9_{\pm 2.2}$ 
& $21.6_{\pm 1.7}$ 
& $43.1_{\pm 0.7}$ \\

EMem \cite{wang2026mem}
& $38.9_{\pm 2.4}$
& $45.4_{\pm 4.2}$
& $43.0_{\pm 2.1}$
& $36.0_{\pm 1.0}$
& $52.4_{\pm 3.1}$
& $31.9_{\pm 2.3}$
& $41.5_{\pm 0.6}$ \\

AMA-Agent \cite{zhao2026ama}
& $\underline{59.1}_{\pm 0.8}$
& $\mathbf{66.9}_{\pm 1.7}$
& $\underline{57.8}_{\pm 1.4}$
& $38.3_{\pm 1.6}$
& $\mathbf{71.7}_{\pm 3.6}$
& $\underline{45.4}_{\pm 1.6}$
& $\underline{56.6}_{\pm 0.8}$ \\

\midrule

\textbf{MESA (ours)}
& $\mathbf{63.2}_{\pm 2.0}$
& $\underline{66.2}_{\pm 2.9}$
& $\mathbf{66.7}_{\pm 2.8}$
& $\mathbf{67.5}_{\pm 3.7}$
& $\underline{71.0}_{\pm 3.4}$
& $\mathbf{53.9}_{\pm 1.7}$
& $\mathbf{65.1}_{\pm 0.8}$ \\

\midrule

\multicolumn{8}{l}{
\textbf{Gemma-4-31B}
} \\
\addlinespace[1pt]

Long-context
& $69.7_{\pm 3.1}$
& $\underline{77.4}_{\pm 0.9}$
& $61.1_{\pm 0.7}$
& $47.3_{\pm 3.1}$
& $63.9_{\pm 2.6}$
& $\mathbf{58.3}_{\pm 1.8}$
& $62.5_{\pm 1.3}$ \\

BM25
& $36.2_{\pm 2.3}$
& $30.7_{\pm 4.1}$
& $21.1_{\pm 2.3}$
& $32.1_{\pm 0.3}$
& $37.0_{\pm 2.2}$
& $16.5_{\pm 1.8}$
& $28.3_{\pm 0.9}$ \\

Qwen3-Emb-4B \cite{zhang2025qwen3}
& $50.7_{\pm 1.2}$
& $59.1_{\pm 1.8}$
& $37.5_{\pm 1.5}$
& $47.8_{\pm 2.1}$
& $50.4_{\pm 2.8}$
& $23.2_{\pm 1.5}$
& $44.2_{\pm 0.7}$ \\

HippoRAG2 \cite{gutierrez2024hipporag}
& $52.5_{\pm 4.2}$
& $46.8_{\pm 2.6}$
& $56.2_{\pm 1.4}$
& $27.2_{\pm 2.0}$
& $66.7_{\pm 3.1}$
& $35.3_{\pm 2.1}$
& $48.2_{\pm 1.3}$ \\

MemGPT \cite{packer2023memgpt}
& $32.7_{\pm 2.4}$
& $39.9_{\pm 1.3}$
& $34.5_{\pm 0.6}$
& $33.3_{\pm 3.1}$
& $49.2_{\pm 3.8}$
& $27.9_{\pm 1.1}$
& $36.2_{\pm 1.4}$ \\

Mem0 \cite{chhikara2025mem0}
& $21.5_{\pm 2.1}$
& $26.0_{\pm 1.9}$
& $17.0_{\pm 1.4}$
& $40.3_{\pm 2.3}$
& $42.7_{\pm 2.5}$
& $9.5_{\pm 1.7}$
& $25.7_{\pm 0.4}$ \\

Hindsight \cite{latimer2025hindsight}
&$50.6_{\pm 3.3}$
&$51.3_{\pm 1.6}$
&$35.8_{\pm 1.5}$
&$41.8_{\pm 2.7}$
&$57.5_{\pm 4.2}$
&$18.2_{\pm 1.4}$
&$42.1_{\pm 1.2}$ \\

A-Mem \cite{xu2026mem}
& $64.0_{\pm 2.6}$
& $60.8_{\pm 2.3}$
& $57.8_{\pm 0.9}$
& $41.8_{\pm 1.9}$
& $45.9_{\pm 2.0}$
& $26.6_{\pm 1.6}$
& $50.0_{\pm 0.9}$ \\

MemoRAG \cite{qian2409memorag}
& $37.5_{\pm 1.0}$
& $42.1_{\pm 2.7}$
& $52.9_{\pm 1.5}$
& $40.3_{\pm 2.7}$
& $\underline{74.8}_{\pm 2.7}$
& $22.8_{\pm 1.4}$
& $46.3_{\pm 0.8}$ \\

EMem \cite{wang2026mem}
& $62.6_{\pm 4.3}$
& $69.7_{\pm 3.0}$
& $61.2_{\pm 0.9}$
& $41.4_{\pm 2.1}$
& $58.7_{\pm 1.7}$
& $53.6_{\pm 1.2}$
& $58.0_{\pm 1.4}$ \\

AMA-Agent \cite{zhao2026ama}
& $\underline{70.3}_{\pm 2.9}$
& $76.9_{\pm 1.6}$
& $\underline{64.7}_{\pm 1.1}$
& $\mathbf{50.9}_{\pm 2.3}$
& $64.8_{\pm 2.4}$
& $50.6_{\pm 2.3}$
& $\underline{63.0}_{\pm 1.0}$ \\

\midrule

\textbf{MESA (ours)}
& $\mathbf{74.3}_{\pm 1.7}$
& $\mathbf{79.4}_{\pm 1.9}$
& $\mathbf{74.6}_{\pm 2.3}$
& $\underline{48.8}_{\pm 2.4}$
& $\mathbf{80.8}_{\pm 2.0}$
& $\underline{55.5}_{\pm 2.1}$
& $\mathbf{69.4}_{\pm 1.1}$ \\

\bottomrule
\end{tabular*}
\caption{
Domain-wise LLM-as-a-judge accuracy (\%) on AMA-Bench using
Qwen3-32B and Gemma-4-31B backbones.
Results are reported as mean $\pm$ sample standard deviation over five splits. 
}
\label{tab:main-ama-domain}
\end{table*}

\subsection{F1 Score on AMA-Bench}
We show the F1 score comparison on AMA-Bench in Table \ref{tab:f1_overall} with two model backbones.

\begin{table}[t]
\centering
\setlength{\tabcolsep}{8pt}
\begin{tabular}{lc}
\toprule
Method & \textbf{F1} \\
\midrule

\multicolumn{2}{l}{\textit{\textbf{LLM Backbone: Qwen3-32B} \cite{yang2025qwen3}}} \\

BM25 & $29.0_{\pm 0.3}$ \\
Qwen3-Emb-4B \cite{zhang2025qwen3} & $28.0_{\pm 0.2}$ \\
MemGPT \cite{packer2023memgpt} & $27.3_{\pm 0.3}$ \\
HippoRAG2 \cite{gutierrez2024hipporag} & $30.6_{\pm 0.4}$ \\
Mem0 \cite{chhikara2025mem0} & $27.6_{\pm 0.3}$ \\
MemoRAG \cite{qian2409memorag} & $31.1_{\pm 0.5}$ \\
Hindsight \cite{latimer2025hindsight} & $28.8_{\pm 0.4}$ \\
A-Mem \cite{xu2026mem} & $\underline{32.7}_{\pm 0.3}$ \\
EMem \cite{wang2026mem} & $31.4_{\pm 0.3}$ \\
AMA-Agent \cite{zhao2026ama} & $31.1_{\pm 0.3}$ \\

\midrule
\textbf{MESA (ours)} & $\mathbf{33.9}_{\pm 0.5}$ \\
\midrule

\multicolumn{2}{l}{\textit{\textbf{LLM Backbone: Gemma-4-31B} \cite{team2026gemma}}} \\

BM25 & $25.8_{\pm 0.1}$ \\
Qwen3-Emb-4B \cite{zhang2025qwen3} & $30.5_{\pm 0.4}$ \\
MemGPT \cite{packer2023memgpt} & $27.4_{\pm 0.5}$ \\
HippoRAG2 \cite{gutierrez2024hipporag} & $32.1_{\pm 0.6}$ \\
Mem0 \cite{chhikara2025mem0} & $22.3_{\pm 0.3}$ \\
MemoRAG \cite{qian2409memorag} & $30.5_{\pm 0.5}$ \\
Hindsight \cite{latimer2025hindsight} & $28.5_{\pm 0.3}$ \\
A-Mem \cite{xu2026mem} & $33.6_{\pm 0.4}$ \\
EMem \cite{wang2026mem} & $35.9_{\pm 0.2}$ \\
AMA-Agent \cite{zhao2026ama} & $\underline{36.0}_{\pm 0.4}$ \\

\midrule
\textbf{MESA (ours)} & $\mathbf{37.4}_{\pm 0.6}$ \\

\bottomrule
\end{tabular}

\caption{
F1 score results of memory-based methods on AMA-Bench using Qwen3-32B and Gemma-4-31B
as answer-model backbones. Results are reported as mean$_{\pm\mathrm{std}}$ over evaluation splits. Best results within each backbone are shown in bold, and second-best results are underlined.
}
\label{tab:f1_overall}
\end{table}

\section{Additional Analyses}
\subsection{Robustness to Judge}
The main AMA-Bench results use Qwen3-32B as the
LLM judge. To avoid the bias of single judge model, we additionally evaluate
the same predictions using Gemma-4-31B and GPT-5.4 in Table~\ref{tab:judge_robustness}. All
three judges receive the same question, reference answer,
predicted answer, and binary correctness prompt.
While the absolute accuracy values vary across judge backbones,
which is consistent with the judge-dependent score variation
also discussed in the original AMA-Agent evaluation
\cite{zhao2026ama}, the relative trend remains stable.
\MESA is ranked first with question-level majority voting under Qwen3-32B, Gemma-4-31B, and
GPT-5.4. This result
indicates that \MESA's comparative advantage is not specific
to the primary Qwen3-32B judge. 
\begin{table}[h]
\centering
\setlength{\tabcolsep}{1pt}
\begin{tabular}{lccc}
\toprule
Method & Qwen3-32B & Gemma-4-31B & GPT-5.4 \\
\midrule
BM25 & $38.5_{\pm 0.6}$ & $36.0_{\pm 0.4}$ & $34.0_{\pm 0.4}$ \\
Qwen3-Emb-4B & $46.6_{\pm 0.4}$ & $39.6_{\pm 0.7}$ & $35.9_{\pm 0.6}$ \\
MemGPT & $33.9_{\pm 0.5}$ & $29.9_{\pm 0.4}$ & $28.5_{\pm 0.5}$ \\
HippoRAG2 & $46.6_{\pm 1.1}$ & $43.4_{\pm 1.0}$ & $39.3_{\pm 1.1}$ \\
Mem0 & $31.1_{\pm 0.6}$ & $26.3_{\pm 0.3}$ & $26.0_{\pm 0.5}$ \\
MemoRAG & $43.1_{\pm 0.7}$ & $39.5_{\pm 0.9}$ & $38.0_{\pm 0.7}$ \\
Hindsight & $36.0_{\pm 0.5}$ & $30.0_{\pm 0.7}$ & $29.5_{\pm 0.6}$ \\
A-Mem & $43.2_{\pm 0.7}$ & $42.3_{\pm 0.5}$ & $39.4_{\pm 1.1}$ \\
EMem & $41.5_{\pm 0.6}$ & $39.4_{\pm 0.9}$ & $39.0_{\pm 0.8}$ \\
AMA-Agent & \underline{56.6}$_{\pm 0.8}$ & \underline{55.2}$_{\pm 0.5}$ & \underline{47.5}$_{\pm 1.1}$ \\
\midrule
\textbf{MESA (ours)} & $\mathbf{65.1}_{\pm 0.8}$ & $\mathbf{59.5}_{\pm 1.8}$ & $\mathbf{52.9}_{\pm 1.4}$ \\
\bottomrule
\end{tabular}
\caption{
Judge Robustness evaluation across different LLM judges (Qwen3-32B, Gemma-4-31B, and GPT-5.4). Overall accuracy (\%) is reported as mean$_{\pm\mathrm{std}}$ over five evaluation splits. Best results within each judge are shown in bold, and second-best results are underlined.
}
\label{tab:judge_robustness}
\end{table}

\subsection{Sensitivity to LoCoMo Conversation Splits}
\label{app:locomo_split_sensitivity}
The main comparison in Table~\ref{tab:locomo} uses a
single fixed conversation-disjoint $2{:}2{:}6$ split shared
by all evaluated methods. Since LoCoMo contains only ten
conversations, results from a single partition may be
sensitive to which conversations are assigned to the
training, validation, and test sets. We therefore additionally
evaluate \MESA over five independently sampled
conversation-level splits using the same partition ratio in Table~\ref{tab:locomo_5fold}.
The fixed split in the main paper remains the basis for the
cross-method comparison, because all baselines are evaluated
under that common protocol. The additional experiments are
intended to characterize the split sensitivity of \MESA.
\begin{table}[h]
\centering
\setlength{\tabcolsep}{8pt}
\begin{tabular}{lc}
\toprule
Method & Score (\%) \\
\midrule
Long-context \cite{yang2025qwen3}& $44.8_{\pm 1.6}$ \\
BM25 & $26.3_{\pm 3.6}$ \\
Qwen3-Emb-4B & $37.4_{\pm 4.6}$ \\ %
MemGPT \cite{packer2023memgpt} & $46.8_{\pm 2.4}$ \\
HippoRAG2 \cite{gutierrez2024hipporag} & $36.6_{\pm 3.1}$ \\ %
Mem0 \cite{chhikara2025mem0} & $37.2_{\pm 5.4}$ \\
MemoRAG \cite{qian2409memorag} & $25.4_{\pm 0.5}$ \\
Hindsight \cite{latimer2025hindsight} & $46.1_{\pm 2.1}$ \\
A-Mem \cite{xu2026mem} & $36.1_{\pm 4.7}$ \\
EMem \cite{wang2026mem} & $\underline{47.0}_{\pm 4.6}$ \\ %
AMA-Agent \cite{zhao2026ama} & $39.7_{\pm 4.8}$ \\
\midrule
\textbf{MESA (ours)} & $\mathbf{49.0}_{\pm 3.3}$ \\ %

\bottomrule
\end{tabular}

\caption{
Performance evaluation over five folds reported as mean$_{\pm\mathrm{std}}$ (\%). Best result is in bold, and second-best result is underlined.
}
\label{tab:locomo_5fold}
\end{table}

\subsection{Memory Construction Time}
We measure offline memory-construction time on a subset of
trajectories sampled from the 208 AMA-Bench episodes and
report the average wall-clock time under a sequential
implementation in Table~\ref{tab:construction_time}. \MESA requires $73.3$s per trajectory,
including $38.2$s for the components shared with
AMA-Agent, $35.1$s for the graph memory,
and a negligible $0.001$s for the raw episodic memory. The memory views
are independently constructed from the same trajectory and
can therefore be parallelized across processes or devices, which needs $38.2$s.

\MESA's construction time is lower than those of Mem0
($179.9$s), Hindsight ($420.3$s), and A-Mem
($717.2$s), while achieving the strongest downstream
accuracy. It is more expensive than AMA-Agent
($38.2$s) and MemoRAG ($32.2$s), reflecting
the additional cost of constructing complementary views.
Overall, the results indicate a acceptable
performance--construction-cost trade-off. 

\begin{table}[h]
\centering
\small
\setlength{\tabcolsep}{1pt}
\renewcommand{\arraystretch}{1.0}
\begin{tabular}{lc}
\toprule
\textbf{Method} & \textbf{Avg. construction time (s)} \\
\midrule
HippoRAG2 \cite{gutierrez2024hipporag} & 35.1 \\
Mem0 \cite{chhikara2025mem0} & 179.9 \\
MemoRAG \cite{qian2409memorag} & 32.2 \\
Hindsight \cite{latimer2025hindsight} & 420.3 \\
A-Mem \cite{xu2026mem} & 717.2 \\
EMem \cite{wang2026mem} & 52.7 \\
AMA-Agent \cite{zhao2026ama} & 38.2 \\
\midrule
\textbf{\MESA (sequential)} & \textbf{73.3} \\
\textbf{\MESA (parallel)} & \textbf{38.2} \\

\bottomrule
\end{tabular}
\caption{Average memory construction time per trajectory (in s) evaluated over sampled episodes with Qwen3-32B.}
\label{tab:construction_time}
\end{table}

\subsection{Comparison with Validation-selected Fixed Selection}
In Sec.~\ref{sec:analysis}, we exhaustively evaluate all non-empty
structure subsets. In Table~\ref{tab:coarse-grained}, we analyze the best fixed combination on the validation set and construct fixed global, domain-conditioned, capability-conditioned, and
domain--capability-conditioned lookup policies. The selected
subsets are then frozen for test evaluation.
Domain-conditioned, capability-conditioned, and domain--capability-conditioned
selection achieve $64.4\%$, $63.8\%$, and $64.0\%$,
respectively. \MESA reaches $65.1\%$ with the lowest
cross-split variation, indicating that its gain is not explained
solely by a coarse category-to-subset mapping. As shown in Fig.~\ref{fig:optimization_trace}, it also learns more fine-grained rules like keyword-level mapping over domain and capacities.
\begin{table}[t]
\centering
\begin{tabular}{lc}
\toprule
\textbf{Selection strategy} & \textbf{Accuracy} \\
\midrule
domain-conditioned &$64.4_{\pm 1.7}$  \\
capability-conditioned& $63.8_{\pm 1.2}$  \\
domain--capability-conditioned & $64.0_{\pm 2.2}$ \\
\textbf{\MESA (ours)} & $\mathbf{65.1}_{\pm 0.8}$ \\
\bottomrule
\end{tabular}
\caption{Comparison with validation-selected coarse-grained selection strategies through the full subset sweep in the controlled analysis.}
\label{tab:coarse-grained}
\end{table}


\section{Discussion and Limitations}
Our results show structure-level selection as a useful complement to memory construction and item-level retrieval. Across AMA-Bench, the best memory configuration is typically neither a single structure nor the full union, indicating that different queries benefit from different combinations of global, temporal, relational, semantic, and step-level evidence. Consistent with this observation, \MESA outperforms the all-structure configuration while using 41.2\% fewer evidence tokens, suggesting that adaptive selection improves both efficiency and evidence quality by reducing redundant or distracting context. The route-to-one ablation further confirms that these gains arise from composing complementary structures rather than identifying a single universally strong representation. Importantly, \MESA learns such selection policies from answer-level feedback without modifying the underlying memory builders, retrievers, or answer model. Prior-guided proposals provide a structured search space, while UCB-guided scheduling improves the stability of policy evolution. The resulting policies also remain explicit and inspectable, facilitating the analysis of query-dependent memory access. The comparatively smaller gain on causal inference points to a promising extension that combines adaptive structure selection with explicit modeling of cross-event and cross-structure dependencies.

This study intentionally focuses on a representative set of five memory structures and keeps the remaining system components fixed to isolate the contribution of query-adaptive selection. This controlled setting enables a clear evaluation of the proposed formulation, while leaving several natural extensions for future work, including incorporating additional memory representations and jointly optimizing memory construction, retrieval, selection, and evidence composition. The current evaluation covers both long-horizon agent trajectories and long-term conversations, providing initial evidence of generality across memory settings. Further studies on continuously updated and multimodal memories could broaden the applicability of the framework.

\end{document}